\documentclass[final,5p,times,twocolumn]{elsarticle}
\usepackage{amsmath}
\usepackage{amssymb}
\usepackage{booktabs}
\usepackage{graphicx}
\usepackage{hyperref}
\usepackage{xcolor}
\usepackage{subfigure}
\usepackage{listings}
\usepackage{longtable}
\usepackage{array}
\usepackage{listings}
\usepackage{multirow}
\usepackage{verbatim}
\usepackage[strings]{underscore}
\usepackage{booktabs}
\usepackage{multirow}
\usepackage{graphicx}
\usepackage{makecell}
\usepackage{tabularx}
\usepackage{float}
\usepackage{threeparttable}

\usepackage{seqsplit}

\newcolumntype{Y}{>{\raggedright\arraybackslash}X}

\usepackage{algorithm}
\usepackage{algpseudocode}

\usepackage{xcolor}
\usepackage{listings}

\definecolor{codebg}{RGB}{248,248,248}
\definecolor{codeframe}{RGB}{210,210,210}
\definecolor{codekeyword}{RGB}{0,76,153}
\definecolor{codestring}{RGB}{160,40,40}
\definecolor{codecomment}{RGB}{90,90,90}

\lstdefinelanguage{json}{
    basicstyle=\ttfamily\small,
    morestring=[b]",
    stringstyle=\color{codestring},
    literate=
     *{0}{{{\color{black}0}}}{1}
      {1}{{{\color{black}1}}}{1}
      {2}{{{\color{black}2}}}{1}
      {3}{{{\color{black}3}}}{1}
      {4}{{{\color{black}4}}}{1}
      {5}{{{\color{black}5}}}{1}
      {6}{{{\color{black}6}}}{1}
      {7}{{{\color{black}7}}}{1}
      {8}{{{\color{black}8}}}{1}
      {9}{{{\color{black}9}}}{1}
      {:}{{{\color{black}:}}}{1}
      {,}{{{\color{black},}}}{1}
      {\{}{{{\color{black}\{}}}{1}
      {\}}{{{\color{black}\}}}}{1}
      {[}{{{\color{black}[}}}{1}
      {]}{{{\color{black}]}}}{1},
}

\lstdefinestyle{paperlisting}{
    backgroundcolor=\color{codebg},
    frame=single,
    rulecolor=\color{codeframe},
    basicstyle=\ttfamily\small,
    breaklines=true,
    numbers=left,
    numberstyle=\tiny\color{codecomment},
    keywordstyle=\color{codekeyword}\bfseries,
    commentstyle=\color{codecomment}\itshape,
    stringstyle=\color{codestring},
    showstringspaces=false,
    captionpos=b,
    tabsize=4
}

\usepackage{xcolor}
\definecolor{codebackground}{RGB}{245,245,245}

\usepackage{makecell}
\usepackage{array}
\newcommand{\yes}{\textbf{Yes}}
\newcommand{\no}{No}
\newcommand{\partialc}{Partial}

\begin{document}

\begin{frontmatter}

\title{DeepUnfolding-Agent: A Classical–Quantum Adaptive Control API}
\author[deakin]{Shanika Iroshi Nanayakkara}
\author[deakin]{Shiva Raj Pokhrel}

\affiliation[deakin]{
    organization={School of Information Technology, Deakin University},
    city={Geelong},
    state={VIC},
    country={Australia}
}





\begin{abstract}
Adaptive optimization workflows in machine learning, federated learning, quantum machine learning, and hybrid classical--quantum systems often embed controller logic directly inside task-specific training scripts. This coupling limits reuse, reproducibility, safety validation, and fair comparison across execution environments. This paper presents \textsc{DeepUnfolding-Agent}, a schema-driven FastAPI platform that exposes deep-unfolded adaptive control as an external software service. External executors register a strategy schema, submit round-level diagnostic variables, receive schema-bounded control actions, execute one optimization round, and return feedback for controller update. The framework separates controller intelligence from domain-specific execution through a strategy registry, diagnostic-to-state adapter, controller factory, safety guard, executor abstraction, feedback-update path, SQLite persistence layer, Docker deployment, and reproducible benchmark scripts. The implementation supports classical, quantum-aware, hybrid, Qiskit Aer, and optional IBM quantum processing execution pathways. We validate the software through unit, schema, API, safety, orchestration, Quantum execution, checkpointing, and Docker tests, and demonstrate external-user fixed-versus-adaptive workflows across finance, healthcare, genomics, transport, noisy-quantum simulation, and hardware-backed IBM Quantum execution. The results show that \textsc{DeepUnfolding-Agent} provides a reusable and traceable adaptive-control layer for heterogeneous optimization workflows. Our quantum experiments are interpreted as API-level quantum-control and hardware-portability evidence.
\end{abstract}


\end{frontmatter}

\section{Introduction}
\begin{figure}[t]
    \centering
    \includegraphics[width=1.0\linewidth]{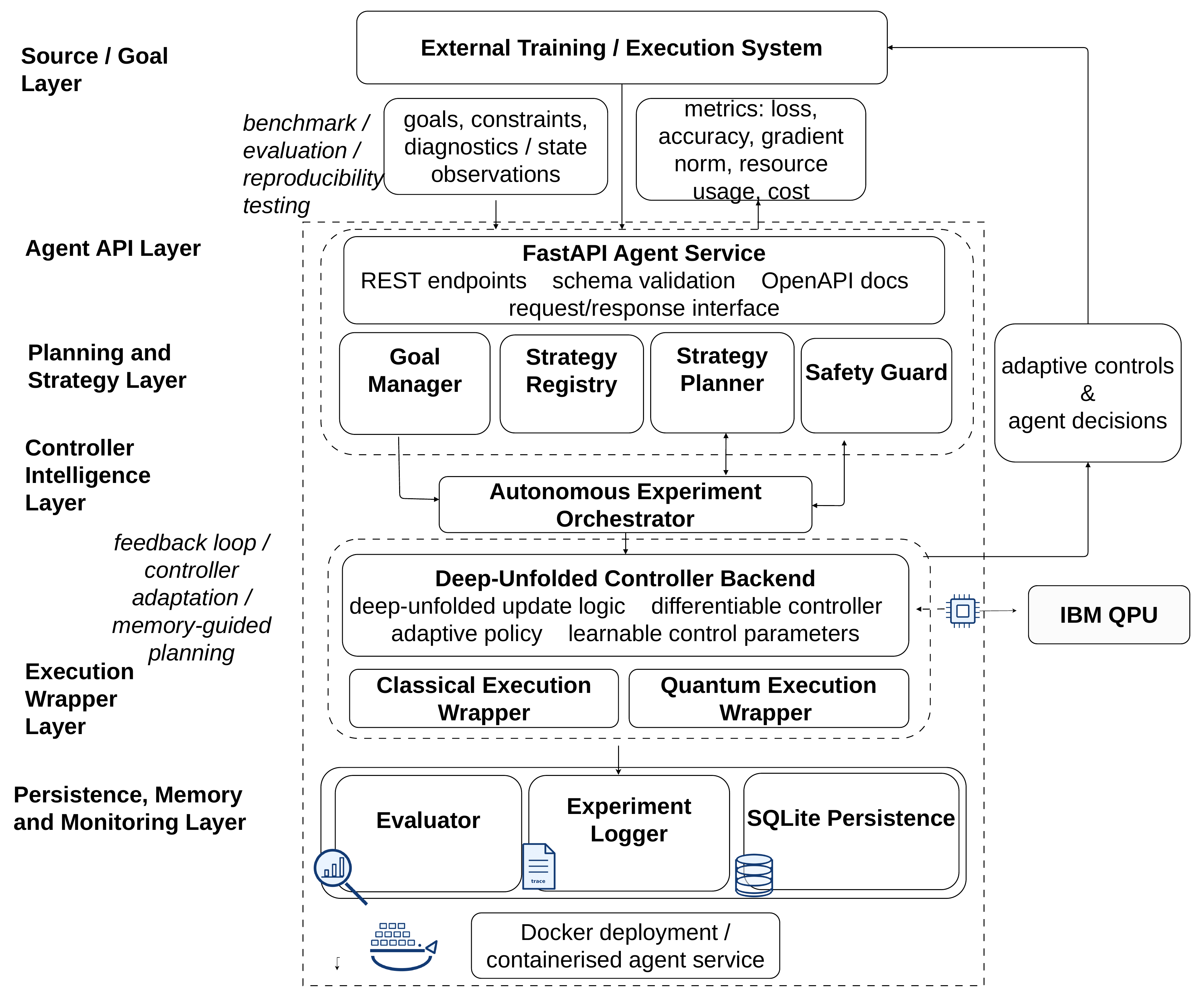}
    \caption{Software architecture of DeepUnfolding-Agent. External workflows retain ownership of data, model training, and backend execution, while the FastAPI service exposes a schema-based controller interface. Runtime diagnostics are validated by the strategy adapter and transformed into controller state vectors. The controller factory instantiates classical, quantum, or hybrid deep-unfolded controllers, which return bounded control variables to the external execution loop. Round-level metrics and feedback are returned to the API for controller update, logging, persistence, and reproducible benchmarking. Optional IBM QPU execution is supported through external Qiskit Runtime scripts using user-owned credentials.}
\label{fig:Software_architecture_highLevel_Overview}
\end{figure}

Iterative optimization is fundamental in modern machine learning (ML), federated learning (FL)~\cite{9079513}, quantum machine learning (QML)~\cite{pokhreldeakin, buyya2024}, and hybrid classical--quantum workflows. In these settings, performance depends not only on the selected learning algorithm but also on training and execution control variables, such as learning rate, local epochs, proximal strength, circuit depth, shot count, and measurement reuse~\cite{kairouz2021advances}. However, these controls often need to adapt to changing runtime diagnostics, including loss, gradient norm, update magnitude, data heterogeneity, shot noise, parameter-shift variance, resource consumption, and convergence behavior~\cite{cerezo2021cost}. Existing adaptive controllers are commonly implemented as task-specific scripts coupled to a particular dataset, model, optimizer, aggregation rule, simulator, or experimental pipeline, making them difficult to reuse, compare, audit, or deploy across external workflows. This coupling reduces reproducibility and limits the practical value of adaptive optimization methods as reusable software components.

\label{sec:abbreviations_notation}
\begin{table}[htb]
\centering
\scriptsize
\caption{List of abbreviations}
\label{tab:abbreviations}
\begin{tabularx}{\linewidth}{p{0.24\linewidth}X}
\toprule
Abbreviation & Meaning \\
\midrule

API & Application Programming Interface \\
DeepUnfolding-Agent & Proposed strategy-agnostic adaptive-control API and software framework \\
DUAgent & DeepUnfolding-Agent, used in figures and tables \\
HTTP & Hypertext Transfer Protocol \\
JSON & JavaScript Object Notation \\
CSV & Comma-Separated Values \\
SQLite & Lightweight relational database used for local persistence \\
Docker & Containerization platform used for reproducible deployment \\
OpenAPI & Machine-readable API specification \\
FL & Federated Learning \\
QFL & Quantum Federated Learning \\
QML & Quantum Machine Learning \\
QPU & Quantum Processing Unit \\
Qiskit Aer & Qiskit simulator backend \\
NISQ & Noisy Intermediate-Scale Quantum \\
Macro-F1 & Unweighted mean of class-wise F1 scores \\
SEM & Standard Error of the Mean \\
pp & Percentage points \\
IID & Independent and identically distributed \\
Non-IID & Non-independent and non-identically distributed \\
BAF & Bank Account Fraud dataset \\
NGSIM & Next Generation Simulation traffic dataset \\
WBC & Wisconsin Breast Cancer dataset \\
\bottomrule
\end{tabularx}
\end{table}

\begin{table}[t]
\centering
\scriptsize
\setlength{\tabcolsep}{3pt}
\renewcommand{\arraystretch}{1.05}
\caption{Core notation used in the API-level control contract.}
\label{tab:core_notation}
\begin{tabularx}{\columnwidth}{@{}p{0.24\columnwidth}X@{}}
\toprule
\textbf{Symbol} & \textbf{Meaning} \\
\midrule
$\mathcal{S}$ & Strategy schema defining diagnostics, controls, bounds, and types \\
$d_t$ & Named diagnostics reported by the external executor at round $t$ \\
$s_t$ & Ordered controller state vector, $s_t=\phi_{\mathcal{S}}(d_t)$ \\
$f_{\theta}^{(K)}$ & Deep-unfolded controller with $K$ unfolding steps \\
$\hat{u}_t$ & Raw controller output before projection \\
$u_t$ & Schema-bounded control vector returned to the executor \\
$E_t$ & External executor applying controls at round $t$ \\
$z_t$ & Execution output after applying $u_t$ \\
$r_t$ & Feedback signal used for controller update \\
$\Pi_{\mathcal{B}_{\mathcal{S}},\mathcal{T}_{\mathcal{S}}}$ & Projection enforcing bounds and data-type constraints \\
\bottomrule
\end{tabularx}
\end{table}

Deep unfolding provides a learnable mechanism for adapting control variables across iterative optimization rounds~\cite{monga2021algorithm,balatsoukas2019deep}. However, many deep-unfolded and adaptive-controller implementations remain embedded within specific algorithms or experimental scripts, limiting reuse because the adaptive logic lacks a stable interface for independent executors. Consequently, comparing classical ML, FL, QML, hybrid optimization, and external domain-specific workflows often requires repeated re-implementation~\cite{shlezinger2025deep, sculley2015hidden}. A reusable software architecture is therefore needed to separate controller intelligence from task execution through a strategy-agnostic, schema-driven interface.

We introduce DeepUnfolding-Agent, abbreviated as DUAgent in figures, tables, and legends, as a strategy-agnostic software framework for adaptive control in classical, quantum workflows. DUAgent exposes deep-unfolded adaptive control through a modular FastAPI service with reusable components for strategy registration, schema adaptation, controller creation, bounded control suggestion, execution routing, feedback collection, controller update, and experiment logging~\cite{fastapi2026,openapi2025,pydantic2026}. Instead of hard-coded training logic, the controller interface is defined by schemas that specify diagnostic state variables, controllable variables, bounds, and data types, enabling a unified API lifecycle for classical, quantum, hybrid, and externally defined strategies.

The primary contribution is an API-level abstraction that separates deep-unfolded controller intelligence from the external execution logic in classical, quantum, and hybrid workflows.
The software is provided as a Docker-deployable FastAPI application with public REST endpoints for health checks, strategy management, controller operations, quantum-controlled execution, autonomous experiment execution, and experiment retrieval~\cite{docker2026}. Validation uses layered software testing, including unit, schema, API endpoint, safety-guard, closed-loop controller, agent orchestration, quantum execution, external executor, and Docker deployment tests. These tests demonstrate that the framework serves as a reusable adaptive-control service for classical, quantum, hybrid, and external-domain workflows.

The main contributions of this paper are threefold:

\begin{enumerate}
\item We present \textsc{DeepUnfolding-Agent}, a schema-driven adaptive-control service that separates controller logic from external optimization workflows. Unlike controllers embedded in training scripts, this service provides controller creation, bounded control suggestions, feedback updates, persistence, and experiment retrieval through a stable FastAPI interface.

\item We introduce a strategy-agnostic diagnostics-controls contract, where each strategy schema specifies observable diagnostic variables, adaptive control variables, data types, and admissible bounds. This approach allows the controller lifecycle to support classical FL, QFL-style workflows, hybrid optimization, and external domain executors without changes to the core API.

\item We implement a bounded closed-loop orchestration layer that handles diagnostic validation, state-vector construction, deep-unfolded control inference, schema and user-bound projection, executor dispatch, feedback-based updates, stopping evaluation, and persistent logging. This ensures traceable and reproducible adaptive-control behavior across multi-round optimization workflows.

\item We validate the software through layered testing and external user evaluations, including unit, schema, API, safety, orchestration, checkpointing, Qiskit Aer, Docker, and optional IBM QPU execution pathways. The experiments confirm API-level re-usability, bounded-control enforcement, reproducible logging, and portability across simulator and hardware-backed quantum execution.
\end{enumerate}

\section{Related Work and Gap}

Existing work relevant to DUAgent spans hyperparameter optimization and AutoML,
deep unfolding, federated-learning frameworks, quantum software frameworks,
MLOps tools and workflow orchestration systems. Table~\ref{tab:sota_comparison}
summarizes the main distinction between these categories and the proposed API-level
adaptive-control framework.

Hyperparameter optimization (HPO) and AutoML tools primarily search over trial-level
configurations, whereas DUAgent operates as a runtime diagnostic--control--feedback
service \cite{agrawal2021hyperparameter,hutter2019automated, akiba2019optuna,liaw2018researchtuna}. Deep-unfolding methods provide learnable iterative structures, but are
usually embedded within a specific model or algorithm \cite{balatsoukas2019deep,monga2021algorithm}.
Federated learning and
Quantum software frameworks provide execution infrastructure, while DUAgent
provides a bounded adaptive-control layer above external execution \cite{mcmahan2017communication,li2020federated,he2020fedml,beutel2020flower, qiskitAer2026}. Therefore, the novelty of the proposed DeepUnfolding-Agent is not only its FastAPI implementation \cite{fastapi2026,openapi2025,pydantic2026}, but its reusable
schema-driven controller lifecycle for classical, quantum, hybrid, and external
optimization workflows.
Further details and a critical review of related work are available in Supplementary Section S2.

\begin{table*}[htb]
\centering
\scriptsize
\caption{Comparison between existing software/method categories and DeepUnfolding-Agent.}
\label{tab:sota_comparison}
\renewcommand{\arraystretch}{1.12}
\setlength{\tabcolsep}{2.5pt}
\resizebox{\textwidth}{!}{%
\begin{tabular}{p{3.1cm}p{3.8cm}ccccc p{4.0cm}}
\toprule
\textbf{Category}
& \textbf{Main capability}
& \makecell{\textbf{Runtime}\\\textbf{control}}
& \makecell{\textbf{Schema}\\\textbf{controls}}
& \makecell{\textbf{Quantum}\\\textbf{aware}}
& \makecell{\textbf{External}\\\textbf{executor}}
& \makecell{\textbf{Bounded}\\\textbf{safety}}
& \textbf{Main gap relative to DUAgent} \\
\midrule

HPO / AutoML~\cite{agrawal2021hyperparameter,hutter2019automated}
& Searches model and hyperparameter configurations.
& \partialc & \no & \no & \partialc & \partialc
& Trial-level search, not round-level diagnostic--control--feedback. \\

Optuna / Ray Tune~\cite{akiba2019optuna,liaw2018researchtuna}
& Supports scalable tuning, pruning, and scheduling.
& \partialc & \no & \no & \partialc & \partialc
& Does not expose a schema-driven controller for every execution round. \\

Deep unfolding~\cite{balatsoukas2019deep,monga2021algorithm}
& Learns parameters of unrolled iterative algorithms.
& \partialc & \no & \partialc & \no & \no
& Usually tied to one algorithm, model, or dataset pipeline. \\

FL frameworks~\cite{mcmahan2017communication,li2020federated,he2020fedml,beutel2020flower}
& Provides distributed FL training and orchestration.
& \partialc & \partialc & \no & \yes & \partialc
& Executes FL but lacks a decoupled quantum-aware adaptive-control API. \\

MLOps tools~\cite{kreuzberger2023machine_mlop}
& Tracks metrics, artefacts, models, and experiments.
& \no & \no & \no & \partialc & \no
& Records experiments but does not generate bounded controls. \\

Quantum frameworks~\cite{qiskitAer2026}
& Supports quantum circuits, simulation, and hardware execution.
& \partialc & \no & \yes & \partialc & \partialc
& Provides execution infrastructure, not cross-domain adaptive control. \\

Workflow/API tools~\cite{fastapi2026,openapi2025,pydantic2026}
& Provides APIs, validation, deployment, and orchestration.
& \partialc & \partialc & \no & \yes & \partialc
& Provides infrastructure but not deep-unfolded control semantics. \\

\textbf{DeepUnfolding-Agent}
& Exposes deep-unfolded adaptive control as a reusable FastAPI service.
& \yes & \yes & \yes & \yes & \yes
& Unified diagnostic--control--feedback API with bounded controls, external executors, controller updates, and reproducible logging. \\

\bottomrule
\end{tabular}%
}
\end{table*}
\section{Design Requirements and Problem Formulation}
\label{sec:design_requirements_problem_formulation}

DeepUnfolding-Agent is designed to address the challenge of reusability in adaptive optimization workflows, where traditional controllers are often tightly integrated with specific datasets, models, or execution environments. This tight coupling makes it difficult to transfer adaptive control logic across different workflows, especially in classical–quantum optimization scenarios that require varying diagnostics, control variables, and backend support.

To overcome these limitations, DeepUnfolding-Agent treats adaptive optimization as a schema-driven control problem. The framework exposes adaptive control as an external, reusable API service, decoupling the controller logic from domain-specific execution.
External executors—responsible for tasks such as data loading, model training, aggregation, or quantum circuit execution—interact with the API by sending runtime diagnostics and receiving bounded control variables.
These controls are applied within the executor's workflow, and the resulting feedback is returned to the API for controller updates. This architecture ensures that adaptive decision-making is clearly separated from task execution, enabling the same controller interface to be reused across diverse classical, quantum, and hybrid optimization workflows without modifying internal training scripts or controller implementations.

This schema-driven approach supports flexible strategy definition, safe bounded control outputs, and compatibility with both classical and quantum workflows, fulfilling the core design requirements for a reusable, extensible, and externally accessible adaptive optimization.

\subsection{Design Requirements}
\label{subsec:design_requirements}

The proposed API is designed according to the following requirements.

\textit{R1: Stable API boundary.}
All interactions with the controller occur through a stable public API, not by importing internal classes. The service provides endpoints for strategy management, controller creation, control suggestion, feedback updates, persistence, and result retrieval.

\textit{R2: Schema-driven strategy definition.}
Optimization strategies are defined by schemas that specify which diagnostic variables the controller observes and which bounded control variables it can adapt. This enables support for both built-in and user-defined workflows across classical, quantum, and hybrid domains

\textit{R3: Strategy and executor-agnostic control.}
The controller operates independently of the external execution logic. It does not require knowledge of the specific optimizer, aggregation rule, simulator, or backend. The executor applies the returned controls, while the API manages diagnostics-to-control adaptation and feedback-driven updates.

\textit{R4: Bounded and safe control outputs.}
All control variables returned to the executor are strictly constrained by schema or user-defined bounds, ensuring only valid and safe values (e.g., learning rates, shot counts, circuit depths) are applied in external workflows.

\textit{R5: Classical and quantum compatibility.}
The API supports both classical and quantum-aware control variables. Classical examples include learning rate, local epochs, and momentum; quantum examples include theta learning rate, shot count, circuit depth, entanglement depth, and measurement reuse.

\textit{R6: Closed-loop feedback adaptation.}
The controller iteratively updates its policy based on feedback metrics (such as validation loss, task score, objective change, stability penalty, or quantum gradient norm) returned by the executor after each round

\textit{R7: Reproducibility and traceability.}
The controller iteratively updates its policy based on feedback metrics (such as validation loss, task score, objective change, stability penalty, or quantum gradient norm) returned by the executor after each round

\textit{R8: External user usability.}
Non-developers can run standard workflows via command-line arguments and auto-generated documentation. Advanced users can extend the system by adding new strategy schemas and executor plugins without modifying the internal API.

\subsection{Problem Formulation and Formal Control Contract}
\label{subsec:problem_formulation_control_contract}

Enable an external optimization workflow to function across communication or control rounds.
For each round $t=1,\ldots, T$, an external executor $E_t$ manages the workflow. The executor may
represent a classical learning pipeline, a federated learning round, a QFL simulation, a noisy quantum execution, or a hardware-backed QPU execution.
The executor is externally owned and manages the dataset, preprocessing, model,
aggregation rule, optimizer, simulator, circuit construction, or hardware
backend. The API does not require direct access to these internal execution
details.

At round $t$, the executor reports a named diagnostic vector.
 $ d_t = \left[d_t^{(1)}, d_t^{(2)}, \ldots, d_t^{(m)}
\right], $
Diagnostics may include validation loss, gradient norm, update norm,
data heterogeneity, shot noise, parameter-shift variance, circuit depth, resource
usage, or round index. 
The meaning and ordering of these diagnostics are defined
by a registered strategy schema, not hard-coded in the controller.

A strategy schema is denoted by 
\begin{equation}
\label{eq:strategy_schema}
\mathcal{S} =
(\mathcal{D}_{\mathcal{S}},
 \mathcal{U}_{\mathcal{S}},
 \mathcal{B}_{\mathcal{S}},
 \mathcal{T}_{\mathcal{S}}),
\end{equation}
where $\mathcal{D}_{\mathcal{S}}$ is the ordered diagnostic set,
$\mathcal{U}_{\mathcal{S}}$ is the ordered control set,
$\mathcal{B}_{\mathcal{S}}$ defines admissible control bounds, and
$\mathcal{T}_{\mathcal{S}}$ defines data-type constraints. For $q$ control
variables, the bounded control domain is
\begin{equation}
\label{eq:bounded_controlDomain}
\mathcal{B}_{\mathcal{S}}
=
\prod_{j=1}^{q}[\ell_j,u_j],
\end{equation}
where $\ell_j$ and $u_j$ are the lower and upper admissible values of the
$j$-th control variable.

The schema adapter maps the named diagnostics into an ordered numerical
controller state:
\begin{equation}
\label{eq:controller_state}
s_t = \phi_{\mathcal{S}}(d_t) \in \mathbb{R}^{m}.
\end{equation}
This mapping validates that all schema-required diagnostics are present and
numeric before the controller is invoked.

The deep-unfolding controller with $K$ unfolding steps is represented as
\begin{equation}
\label{eq:deepunfolding_controller}
f_{\theta}^{(K)}:\mathbb{R}^{m}\rightarrow\mathbb{R}^{q},
\end{equation}
where $\theta$ denotes the trainable controller parameters. Given the state
vector $s_t$, the controller generates raw control suggestions
\begin{equation}
\label{eq:raw_controlsuggestion}
   \hat{u}_t = f_{\theta}^{(K)}(s_t). 
\end{equation}

These raw controls are not returned directly to the executor. They are first
projected into the schema-defined feasible set:
\begin{equation}
\label{eq:feasibleSet}
u_t =
\Pi_{\mathcal{B}_{\mathcal{S}},\mathcal{T}_{\mathcal{S}}}
(\hat{u}_t),
\end{equation}
where $\Pi_{\mathcal{B}_{\mathcal{S}},\mathcal{T}_{\mathcal{S}}}$ denotes
clipping, type conversion, and safety validation according to the schema bounds,
data-type constraints, and any user-defined safety limits. The bounded vector
$u_t$ is the final control vector returned by the API and applied by the
external executor.

The executor then performs one optimization round using the bounded controls:
\begin{equation}
\label{eq:executor_output}
z_t = E_t(u_t),
\end{equation}
where $z_t$ denotes round-level execution outputs such as training loss,
validation loss, task score, update norm, quantum expectation value, shot usage,
circuit depth, or other workflow-specific metrics. A feedback mapping extracts
the controller-update signal from these outputs:
$r_t = \psi(z_t).$
The feedback $r_t$ may include objective improvement, validation loss, stability
penalty, resource penalty, task score, or quantum-specific metrics.

The controller is updated using the observed state, bounded controls, and
feedback:
\begin{equation}
\label{eq:updatestates}
\theta_{t+1}
=
\mathcal{A}(\theta_t, s_t, u_t, r_t),
\end{equation}
where $\mathcal{A}$ denotes the feedback-based controller update rule. This
update allows the controller to adapt its control policy over repeated execution
rounds.

The overall control objective is to select schema-valid controls that improve
the external workflow objective while satisfying all bounds:
\begin{equation}
\label{eq:objective}
\min_{\theta}
\sum_{t=1}^{T}
\mathcal{J}_t(z_t),
\end{equation}
subject to
$
u_t \in \mathcal{B}_{\mathcal{S}}, \qquad t=1,\ldots,T,$
with
\begin{equation}
\label{eq:bounds}
z_t =
E_t\left(
\Pi_{\mathcal{B}_{\mathcal{S}},\mathcal{T}_{\mathcal{S}}}
\left(
f_{\theta}^{(K)}(\phi_{\mathcal{S}}(d_t))
\right)
\right).
\end{equation}
Here, $\mathcal{J}_t$ represents the round-level objective or cost induced by the
external workflow, such as validation loss, convergence error, instability,
resource cost, or a weighted combination of these terms.

This formulation defines the API-level control contract used throughout the
proposed framework: diagnostics are mapped to a schema-ordered state vector, the
controller generates raw controls, projection enforces bounds, and type
constraints, the external executor applies the bounded controls, and feedback is returned for controller update. Consequently, the same formulation applies to
classical, quantum, hybrid, and user-defined optimization workflows.

\subsection{Classical and Quantum Instantiations}
\label{subsec:classical_quantum_instantiations}

The formal control contract can be instantiated with different strategy schema. 
In a classical federated learning workflow, the diagnostic vector may be written as
$
d_t^{\mathrm{classical}}
=
[
\ell_{t-1},
g_t,
\Delta_t,
h_t,
t
],$
where \(\ell_{t-1}\) denotes previous loss, \(g_t\) denotes gradient norm,
\(\Delta_t\) denotes update norm, \(h_t\) denotes data heterogeneity, and \(t\)
denotes the round index.
The corresponding bounded classical control vector may be written as
$
u_t^{\mathrm{classical}}
=
[
\alpha_t,
E_t^{loc},
\mu_t,
\beta_t,
\lambda_t
],$
where \(\alpha_t\) is the learning rate, \(E_t^{loc}\) is the number of local epochs,
\(\mu_t\) is the proximal strength, \(\beta_t\) is momentum, and \(\lambda_t\) is
weight decay.
The external executor applies these controls inside the selected aggregation strategy and optimizer, such as FedAvg, FedProx, FedNova, SGD, Adam, AdamW, or a user-defined alternative. The API does not implement the aggregation rule or optimizer internally; it only suggests bounded controls according to the registered schema and updates the controller based on the returned feedback.

In a quantum federated learning workflow, the diagnostic vector may be written as
$
d_t^{\mathrm{qfl}}
=
[
\ell_{t-1},
g_t^{q},
v_t^{ps},
\eta_t^{shot},
c_t,
t
],$
where \(\ell_{t-1}\) denotes previous loss, \(g_t^{q}\) denotes quantum gradient norm,
\(v_t^{ps}\) denotes parameter-shift variance, \(\eta_t^{shot}\) denotes shot noise,
\(c_t\) denotes circuit depth, and \(t\) denotes the round index.
The corresponding bounded quantum control vector may be written as
$
u_t^{\mathrm{qfl}}
=
[
\alpha_t^{q},
N_t^{shot},
c_t^{ctrl},
e_t,
\rho_t
],$ where \(\alpha_t^{q}\) is the theta learning rate, \(N_t^{shot}\) is the shot count,
\(c_t^{ctrl}\) is the controlled circuit depth, \(e_t\) is the entanglement depth,
and \(\rho_t\) is the measurement-reuse factor.
The external quantum executor applies these controls using a QFL simulator, Qiskit Aer, noisy quantum stress tests, or hardware-backed IBM QPU workflows. Backend-specific tasks, including circuit construction, simulation, transpilation, runtime execution, and job submission, are managed within the executor.

The same abstract control formulation supports both classical and quantum optimization. Only the strategy schema and external executor change; the API lifecycle remains consistent: create a controller, send diagnostics, receive bounded controls, execute externally, return feedback, update the controller, and persist evidence. This approach informs the software architecture described in the next section.

\section{DeepUnfolding-Agent Architecture}
\label{sec:duagent_architecture}
\begin{figure}[htb]
    \centering
\includegraphics[width=0.9\linewidth]{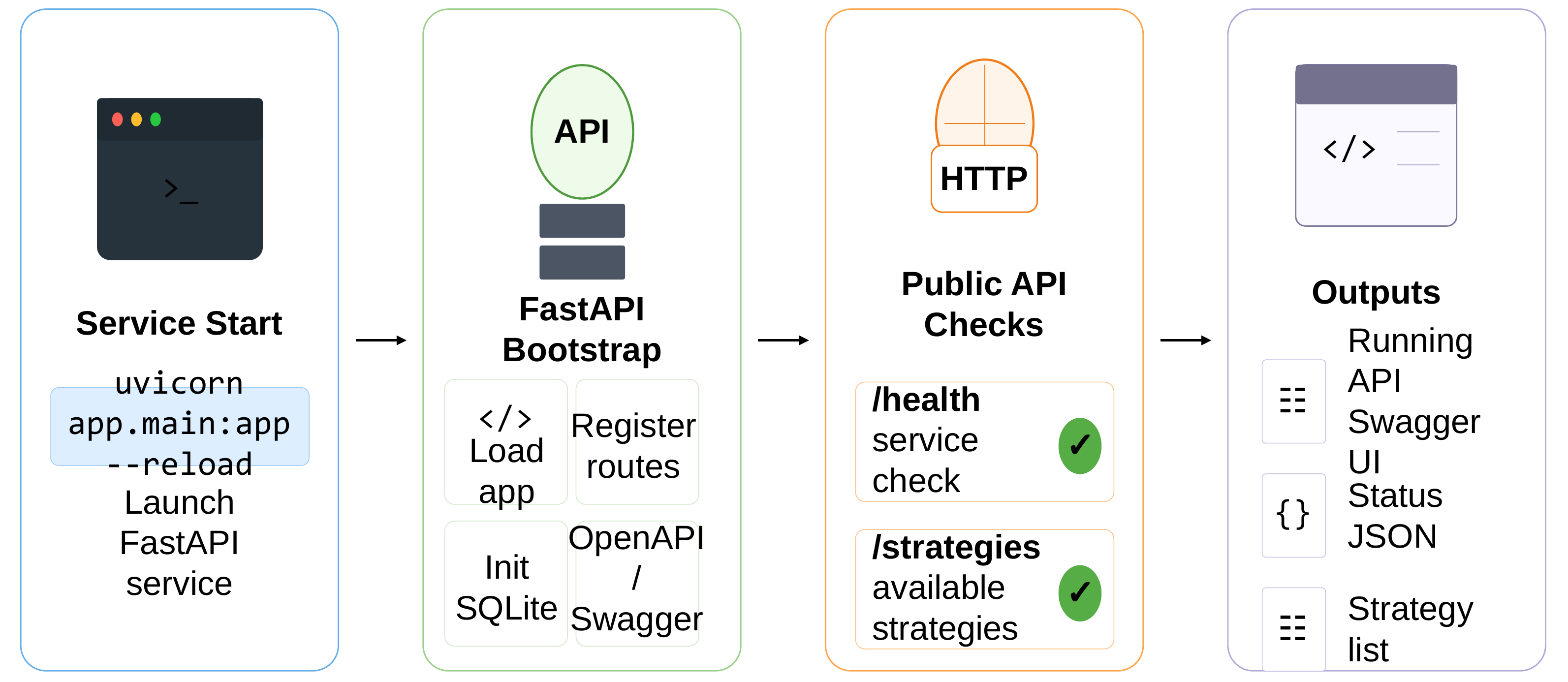}
    \caption{Service Entry API Discovery}
\label{fig:Service_Entry_API_Discovery}
\end{figure}

\begin{figure*}[htb]
    \centering
    \includegraphics[width=0.8\linewidth]{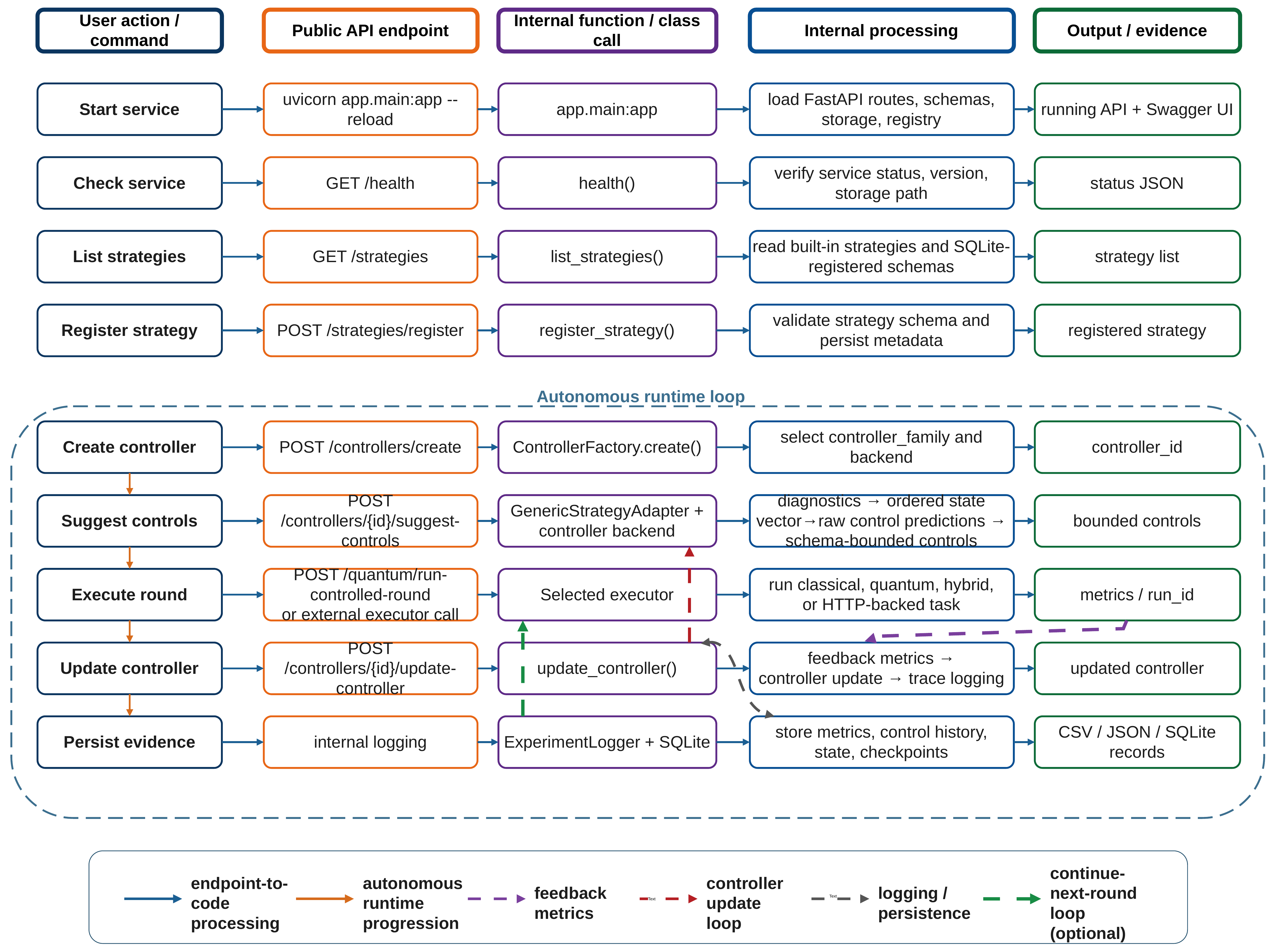}
    \caption{API-to-code execution workflow of DeepUnfolding-Agent. 
The figure maps external user commands and public API endpoints to the corresponding internal functions, processing stages, and output artifacts. 
The upper part shows service initialization, health checking, strategy listing, and user-defined strategy registration. 
The dashed region highlights the autonomous runtime loop, where a controller is created, diagnostics are converted into bounded control suggestions, an execution round is performed, feedback metrics are returned, the controller is updated, and evidence is persisted as CSV, JSON, and SQLite records. 
Different arrow styles indicate endpoint-to-code processing, autonomous runtime progression, feedback flow, controller-update flow, logging/persistence, and optional continuation to the next round. Specifically, Supplementary Figures~S1--S6 decompose the main API-to-code
workflow into its internal software stages.}
\label{fig:api_to_code_workflow}
\end{figure*}

DeepUnfolding-Agent is a schema-driven adaptive-control service. Its architecture implements the formal control contract described in Section~\ref{subsec:problem_formulation_control_contract}. Within this framework, named diagnostics are converted into an ordered state vector, a deep-unfolded controller generates raw control signals, schema projection enforces safety and data consistency, an external executor applies bounded controls, and feedback supports controller updates and persistence. This separation of adaptive decision-making from domain-specific execution allows the controller lifecycle to be reused across classical, quantum, or even hybrid, and user-defined optimization workflows.

The service is available through a FastAPI interface. Public endpoints enable strategy discovery and registration, controller creation and updates, control suggestions, quantum execution, experiment retrieval, and persistence. Internally, the architecture consists of six main components: strategy schema, diagnostic adapter, deep-unfolded controller backend, safety-bounded projection layer, external executor protocol, and persistence layer.

\begin{algorithm}[htb]
\caption{Schema-Bounded DeepUnfolding-Agent Control Loop}
\label{alg:duagent}
\begin{algorithmic}[1]
\Require Strategy schema $\mathcal{S}$, external executor $E$, deep-unfolded controller $f_{\theta}^{(K)}$, maximum rounds $T$
\State Register or retrieve strategy schema $\mathcal{S}$
\State Create controller instance $c$ using $\mathcal{S}$
\State initialize run logger and controller metadata store
\For{$t=1,\ldots,T$}
    \State Receive runtime diagnostics $d_t$ from executor $E$
    \State Construct ordered state vector $s_t=\phi_{\mathcal{S}}(d_t)$
    \State Generate raw controls $\hat{u}_t=f_{\theta}^{(K)}(s_t)$
    \State Apply schema and user safety bounds $u_t=\Pi_{\mathcal{B}_{\mathcal{S}}}(\hat{u}_t)$
    \State Execute one external optimization round $y_t=E(u_t)$
    \State Compute feedback $r_t$ from validation, stability, and resource metrics in $y_t$
    \State Update controller instance $c$ using feedback $r_t$
    \State Persist diagnostics $d_t$, state vector $s_t$, controls $u_t$, feedback $r_t$, metrics $y_t$, and controller metadata
    \If{stopping criterion is satisfied}
        \State \textbf{break}
    \EndIf
\EndFor
\State \Return run summary, control trace, metrics, and persisted artifacts
\end{algorithmic}
\end{algorithm}

\begin{figure*}[htb]
    \centering
    \includegraphics[width=0.85\linewidth]{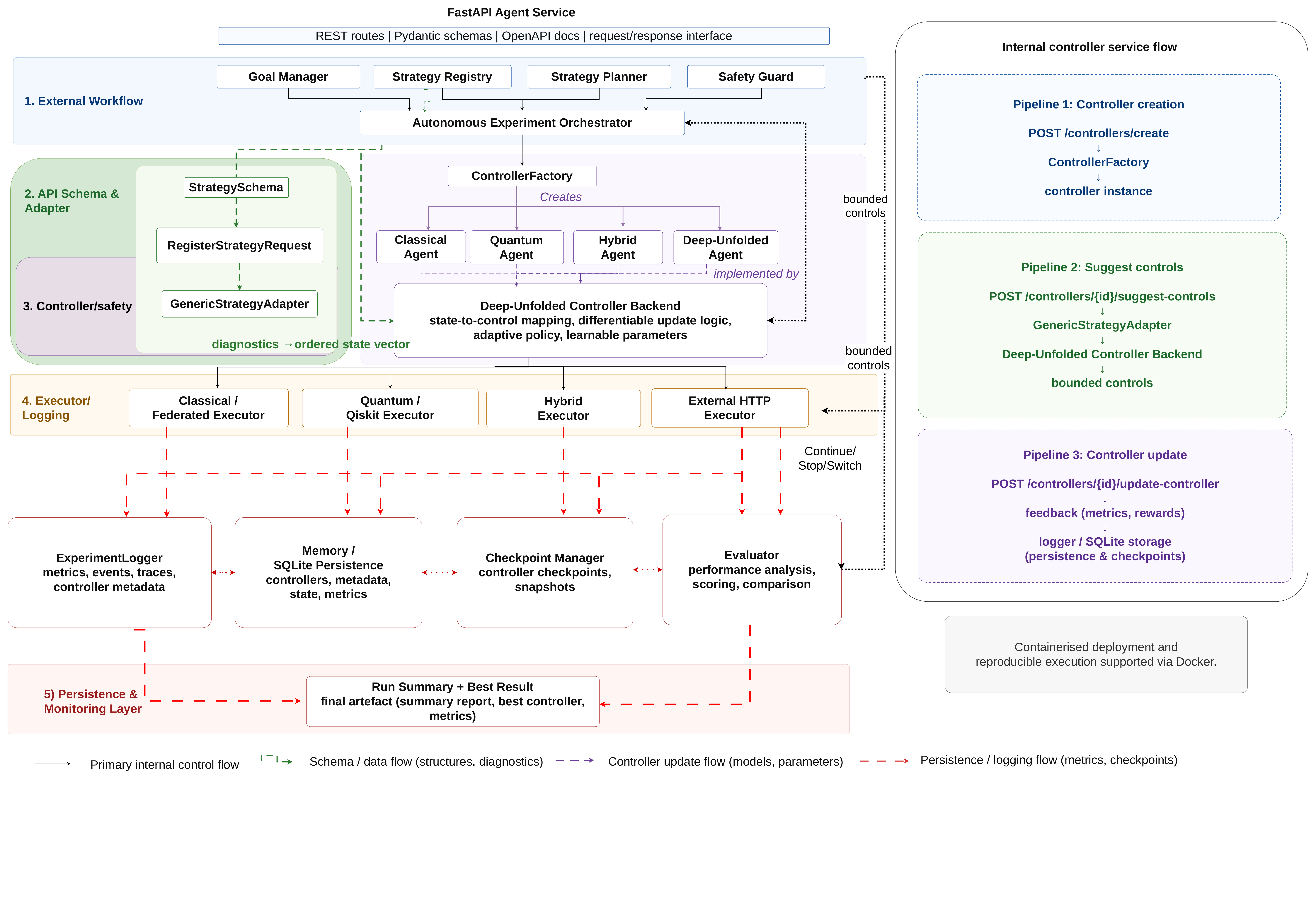}
    \caption{Internal Software Architecture of the Autonomous Agent Controller}
    \label{fig:placeholder}
\end{figure*}

Figure~\ref{fig:api_to_code_workflow} presents the complete API-to-code execution workflow.
The following subsections break down this workflow into service entry point, strategy registration, agent creation, control suggestion, execution, feedback update, and persistence stages.

This stage corresponds to the upper part of Figure~\ref{fig:api_to_code_workflow}, where user-level commands are mapped to public API endpoints and internal route handlers. Its role is to establish the public software boundary: users interact with endpoint calls, while the internal implementation manages routing, schema validation, registry access, and persistent storage.

Detailed component-level pipeline diagrams are provided in the Supplementary Section~S3.

The service entry and API discovery process is performed as illustrated in Figure \ref{fig:Service_Entry_API_Discovery}.

\subsection{Strategy Schema}
\label{subsec:architecture_strategy_schema}

The strategy schema serves as the core abstraction that keeps the API independent of any specific optimizer, aggregation rule, quantum circuit, simulator, or dataset. Instead of hard-coding controller inputs and outputs, each workflow is defined by a registered schema. As shown in Equation~\ref{eq:strategy_schema}, the schema specifies the ordered diagnostic variables, ordered control variables, admissible bounds, and data-type constraints.

The \texttt{/strategies/register} endpoint implements this abstraction in software. Users can register either a built-in or user-defined strategy by specifying the diagnostic variables observed by the controller and the control variables that may be adapted. The API validates the strategy identifier, controller family, diagnostic list, control list, bounds, and data types before storing the schema in SQLite. Once stored, the schema can be reused for future controller-creation requests without changes to the public API, controller factory, or controller backend.

This design accommodates both built-in and user-defined workflows. For example, a classical FL schema may define diagnostics such as loss, gradient norm, update norm, and data heterogeneity, while a QFL schema may specify quantum diagnostics such as shot noise, parameter-shift variance, and circuit depth. The API does not require knowledge of the internal meaning of the workflow; it only needs the schema-defined diagnostic and control interface.

\subsection{Diagnostic Adapter}
\label{subsec:architecture_diagnostic_adapter}

The diagnostic adapter forms the boundary between external runtime observations and the internal controller representation. External executors report diagnostics as named values, such as \texttt{previous\_loss}, \texttt{gradient\_norm}, \texttt{shot\_noise}, or \texttt{circuit\_depth}. The controller backend, however, requires an ordered numerical state vector. The adapter therefore implements the mapping defined in Equation~\ref{eq:controller_state}.

The adapter applies the diagnostic order from the selected strategy schema. It verifies that all required diagnostics are present and numeric, then converts them into the ordered state vector \(s_t\). If any diagnostic is missing or invalid, the request is rejected before reaching the controller. This approach prevents state-vector corruption and enhances API safety for external users.

Since the adapter relies solely on the registered schema, it supports classical, quantum, hybrid, and user-defined workflows. This strategy-agnostic bridge allows external systems to use domain-specific diagnostic names, while the controller receives a consistent numerical representation.

\subsubsection{Validation and Error-Handling Policy}
\label{subsubsec:diagnostic_error_handling}

The diagnostic adapter uses a fail-fast validation policy before controller inference. For each selected strategy schema, all required diagnostic variables in \(\mathcal{D}_{\mathcal{S}}\) must be present in the request. If any are missing, the request is rejected and no controller suggestion is generated. Each diagnostic value must be convertible to a numeric value; non-numeric diagnostics are rejected before constructing the state vector. This approach prevents silent corruption of the ordered controller state \(s_t\) and ensures the deep-unfolded backend receives only schema-compatible numerical inputs.

The current implementation enforces a strict policy: required diagnostics do not receive implicit defaults within the API adapter. While external demonstration scripts may use defaults when creating initial diagnostics, all diagnostics submitted to the API must comply with the registered strategy schema. This design exposes integration errors to external users instead of masking them through automatic imputation.

Out-of-range diagnostic values are handled differently from out-of-range controls. The current schema mainly defines bounds for control variables, so diagnostic values are only checked for presence and numeric type and are not typically clipped by the API. If bounded diagnostics are needed, bounds can be added to the strategy schema or enforced by the external executor before submission.

\subsection{Deep-Unfolded Controller Backend}
\label{subsec:deep_unfolded_controller_backend}

Once a strategy is selected, the user creates a controller instance using the \texttt{/controllers/create} endpoint. This endpoint retrieves the relevant strategy schema from either the built-in or SQLite-backed user registry. The schema is provided to the controller factory, which determines the appropriate controller family and backend. The controller instance is stored in active process memory, and its metadata is saved for future inspection, saving, loading, and experiment tracing.

The deep-unfolded controller backend applies the control policy defined in Equation~\ref{eq:deepunfolding_controller}. Given a schema-ordered state vector, the controller generates raw control suggestions as described in Equation~\ref{eq:raw_controlsuggestion}. These suggestions may include classical controls, such as learning rate, local epochs, proximal strength, momentum, or weight decay, as well as quantum-aware controls, such as theta learning rate, shot count, circuit depth, entanglement depth, or measurement reuse.

This design separates controller construction from usage. External users do not instantiate internal Python classes directly. Instead, they create a reusable controller via a stable API endpoint and use the returned \texttt{controller\_id} for subsequent control-suggestion, execution, update, save, and load operations.

\subsection{Safety-Bounded Projection}
\label{subsec:safety_bounded_projection}

The safety-bounded projection layer implements Equation~\ref{eq:feasibleSet}. Rather than returning raw controller outputs to the external executor, it projects each suggested control into the admissible range defined by the strategy schema and user-defined safety constraints. This process includes clipping, data-type conversion, and control value validation.

This layer is essential because external workflows often have strict operational limits. Classical workflows may require learning rates, local epochs, regularization strengths, or optimizer parameters to stay within valid ranges. Quantum workflows may restrict shot count, circuit depth, entanglement depth, measurement reuse, backend budget, or hardware-supported execution parameters. The projection layer ensures the executor receives only bounded and valid controls.

\subsubsection{Safety-Bounded Control Suggestion}
\label{subsec:safety_bounded_control_suggestion}

The control-suggestion stage implements the main inference path of the adaptive controller. Given a controller identifier and a diagnostics object, the API first retrieves the corresponding controller instance and registered strategy schema. The diagnostic adapter then converts the named diagnostics into the ordered controller state \(s_t\). The deep-unfolded controller maps this state to raw control suggestions, which are then projected into the schema-defined feasible set before being returned to the executor.

Formally, the bounded control vector returned at round \(t\) is
\begin{equation}
\label{eq:bounded_control_suggestion}
u_t =
\Pi_{\mathcal{B}_{\mathcal{S}},\mathcal{T}_{\mathcal{S}}}
\left(
f_{\theta}^{(K)}(s_t)
\right).
\end{equation}
Here, \(f_{\theta}^{(K)}\) denotes the deep-unfolding controller with \(K\) unfolding steps, \(s_t\) is the schema-ordered diagnostic state vector, and \(\Pi_{\mathcal{B}_{\mathcal{S}}}\) is the projection operator that enforces the admissible bounds defined by the strategy schema. In practice, this projection clips continuous controls to their specified bounds, converts integer-valued controls to valid integers, and rejects or corrects invalid outputs based on the schema and user-defined safety constraints.

This projection step is essential because raw controller outputs may not be directly valid for the external workflow. In classical workflows, it prevents invalid learning rates, epoch counts, proximal strengths, momentum values, or weight-decay parameters. In quantum workflows, it constrains shot counts, circuit depth, entanglement depth, measurement reuse, and quantum learning-rate values to ranges supported by the simulator or hardware backend. As a result, the executor receives only bounded controls \(u_t\), not the raw controller output.

The suggested controls are logged along with the original diagnostics, constructed state vector, controller identifier, and strategy identifier. This ensures traceability between the external runtime state, controller decisions, projected bounded controls, and subsequent feedback updates.

The ${/controllers/\{controller\_id\}/suggest-controls} $ endpoint performs three tasks: it converts diagnostics into a state vector, retrieves raw controls from the controller backend, and returns only schema-bounded controls to the external workflow. Returned controls, diagnostics, and the state vector are logged to ensure traceability.

\subsubsection{Control Defaults and Bound Enforcement}
\label{subsubsec:control_defaults_bounds}

Control values are managed using a bounded projection policy. The executor does not apply raw controller outputs \(\hat{u}_t\) directly. Instead, outputs are converted to the required data type and clipped to the admissible range defined by \(\mathcal{B}_{\mathcal{S}}\) and \(\mathcal{T}_{\mathcal{S}}\). This process ensures the final control vector \(u_t\) contains only values valid for the chosen classical or quantum workflow.

For external-user runners, fallback defaults are applied only if a control key is missing from the controller response. For example, the classical runner may use a conservative learning rate, zero proximal strength, and one local epoch, while the QFL runner may use conservative defaults for theta learning rate, shot count, circuit depth, entanglement depth, and measurement-reuse value. These defaults are clipped to the configured bounds before execution. Defaults serve as an executor-side robustness measure, while schema validation remains the API-side contract.

\begin{table}[htb]
\centering
\scriptsize
\caption{Validation and error-handling policy used by DeepUnfolding-Agent.}
\label{tab:error_handling_policy}
\begin{tabular}{p{0.28\linewidth}p{0.30\linewidth}p{0.32\linewidth}}
\toprule
Case & Policy & Location \\
\midrule
Missing required diagnostic
& Reject request before controller inference
& Diagnostic adapter \\
Non-numeric diagnostic
& Reject request before state-vector construction
& Diagnostic adapter \\
Out-of-range diagnostic
& Not clipped by default; may be constrained by executor or extended schema
& Executor/schema extension \\
Missing control key
& Use conservative fallback default, then clip
& External runner/projection layer \\
Out-of-range control
& Clip to schema or user-defined bounds
& Safety-bounded projection \\
Invalid control type
& Convert to required type where possible; otherwise reject or fail execution
& Projection layer/executor normalization \\
Missing controller identifier
& Return controller-not-found error
& API controller registry \\
Unsupported strategy
& Register user schema or return unknown-strategy error
& Strategy registry \\
\bottomrule
\end{tabular}
\end{table}

\subsection{External Executor Protocol}
\label{subsec:external_executor_protocol_architecture}

The external executor protocol implements the execution step in Equation~\ref{eq:executor_output}. The API does not own the dataset, training loop, aggregation strategy, optimizer, simulator, circuit construction, or QPU job submission. These remain inside the external executor. The API only supplies bounded controls and receives feedback after execution.

In classical workflows, the executor applies bounded controls during federated learning rounds, local optimizer updates, or user-defined training loops. In quantum workflows, it applies controls within a QFL simulator, a Qiskit Aer execution path, a noisy quantum stress test, or an IBM QPU workflow. This approach keeps backend-specific execution details separate from the controller service while maintaining a consistent diagnostic, control, and feedback interface.

The controlled execution endpoint demonstrates this protocol for quantum execution. It retrieves the controller and schema, converts diagnostics to a state vector, obtains bounded controls, sends these controls to the selected quantum executor, collects execution metrics, and prepares feedback for controller updates. External classical and hybrid workflows follow the same interaction pattern.

\subsection{Feedback Update, Persistence, and Reproducibility}
\label{subsec:feedback_persistence_reproducibility}

The feedback-update stage converts execution metrics into a controller-update signal. After the external executor applies the bounded controls \(u_t\), it returns round-level outputs \(z_t\), such as validation performance, stability indicators, resource usage, and backend-specific execution metadata. These outputs are then mapped to a scalar feedback objective and, optionally, a structured feedback object.

We define the feedback objective at round \(t\) as
\begin{equation}
\label{eq:feedback_objective}
\mathcal{F}_t
=
w_{\mathrm{val}}\mathcal{L}^{\mathrm{val}}_t
+
w_{\mathrm{stab}}\Omega^{\mathrm{stab}}_t
+
w_{\mathrm{res}}\Omega^{\mathrm{res}}_t
+
w_{\mathrm{lat}}\Omega^{\mathrm{lat}}_t,
\end{equation}
where \(\mathcal{L}^{\mathrm{val}}_t\) is the validation loss, 
\(\Omega^{\mathrm{stab}}_t\) is a stability penalty, 
\(\Omega^{\mathrm{res}}_t\) is a resource penalty, and 
\(\Omega^{\mathrm{lat}}_t\) is a latency or execution-cost penalty. 
The weights \(w_{\mathrm{val}}, w_{\mathrm{stab}}, w_{\mathrm{res}}, w_{\mathrm{lat}}\geq 0\) control the relative importance of prediction quality, stable optimization, resource efficiency, and execution cost.

The stability penalty captures unstable optimization behavior, such as large update norms, high gradient norms, oscillatory validation loss, or quantum-gradient instability. A typical classical stability term may be written as
\begin{equation}
\label{eq:stability_penalty}
\Omega^{\mathrm{stab}}_t
=
\| \Delta \theta_t \|_2
+
\gamma_g \| g_t \|_2,
\end{equation}
where \(\Delta \theta_t\) is the model update, \(g_t\) is the gradient estimate, and \(\gamma_g\) is a weighting coefficient. In QFL workflows, this term can also include quantum-gradient norm or parameter-shift variance.

The resource penalty represents the cost of the selected controls. For quantum execution, this may include shot count and circuit depth:
\begin{equation}
\label{eq:quantum_resource_penalty}
\Omega^{\mathrm{res}}_t
=
\gamma_N
\frac{N^{\mathrm{shot}}_t}{N^{\mathrm{shot}}_{\max}}
+
\gamma_c
\frac{c_t}{c_{\max}},
\end{equation}
where \(N^{\mathrm{shot}}_t\) is the number of shots, \(c_t\) is the circuit depth, and \(N^{\mathrm{shot}}_{\max}\) and \(c_{\max}\) are the maximum allowed shot count and circuit depth. For classical workflows, the resource penalty may instead represent local epochs, communication cost, memory use, or compute time.

The latency penalty captures execution delay or backend cost:
\begin{equation}
\label{eq:latency_penalty}
\Omega^{\mathrm{lat}}_t
=
\frac{\tau_t}{\tau_{\max}},
\end{equation}
where \(\tau_t\) is the measured or estimated execution latency and \(\tau_{\max}\) is a normalising constant.

The structured feedback object returned to the controller can therefore be written as
$r_t =
\left\{
\mathcal{L}^{\mathrm{val}}_t,
\Delta \mathcal{J}_t,
\Omega^{\mathrm{stab}}_t,
\Omega^{\mathrm{res}}_t,
\Omega^{\mathrm{lat}}_t,
\mathcal{F}_t
\right\}, $
where \(\Delta \mathcal{J}_t\) denotes the observed objective improvement or degradation relative to the previous round. The controller update is then expressed as
$
\theta_{t+1}
=
\mathcal{A}(\theta_t, s_t, u_t, r_t).$

This approach clarifies the feedback update process: the controller is updated based on a multi-objective feedback signal, not solely on final task accuracy. This signal can include validation quality, optimization stability, resource usage, and execution cost.

In the current implementation, metrics such as validation loss, objective change, stability penalty, update norm, expectation value, shot count, and circuit-depth metadata are logged based on the selected workflow. Resource and latency penalties are extensible feedback terms that external executors can compute and submit through the same feedback object, without requiring changes to the API contract.

\section{API Lifecycle and External-Executor Contract}
\label{sec:api_design_usage}

DeepUnfolding-Agent exposes adaptive control through a stable REST-based API
rather than requiring external users to import internal Python classes. The API
defines a diagnostic--control--feedback lifecycle in which an external executor
registers or selects a strategy schema, creates a controller, and submits the runtime
diagnostics, receives bounded controls, executes one optimization round, returns
feedback, and retrieves logged experiment artifacts. This interface keeps
domain-specific training and execution code outside the controller service, while
the API manages schema validation, controller inference, safety-bounded projection,
feedback updates, and persistence.

The API lifecycle is organized around four user-facing stages. First, a strategy
schema defines the ordered diagnostic variables, control variables, bounds, and
data types used by a workflow. Second, a controller instance is created for the
selected schema and assigned a reusable controller identifier. Third, external
executors call the control-suggestion endpoint at each round to convert
schema-compatible diagnostics into bounded controls. Finally, execution feedback
is returned to update the controller and persist the run trace. This lifecycle is
consistent across classical, quantum-aware, hybrid, and user-defined workflows.

\begin{table}[t]
\centering
\scriptsize
\setlength{\tabcolsep}{2.5pt}
\renewcommand{\arraystretch}{1.12}
\caption{Compact public API lifecycle exposed by DeepUnfolding-Agent.}
\label{tab:compact_api_lifecycle}
\begin{tabular}{@{}p{0.36\columnwidth}p{0.40\columnwidth}p{0.18\columnwidth}@{}}
\toprule
\textbf{Endpoint} & \textbf{Purpose} & \textbf{Role} \\
\midrule

\texttt{/health}
& Checks service status, version, and storage
& Inspection \\

\texttt{/strategies}
& Lists built-in and registered schemas
& Discovery \\

\makecell[l]{\texttt{/strategies/}\\\texttt{register}}
& Registers a strategy schema
& Extensibility \\

\makecell[l]{\texttt{/controllers/}\\\texttt{create}}
& Creates a controller instance
& Creation \\

\makecell[l]{\texttt{/controllers/\{id\}/}\\\texttt{suggest-controls}}
& Returns bounded controls from diagnostics
& Inference \\

\makecell[l]{\texttt{/controllers/\{id\}/}\\\texttt{update-controller}}
& Updates the controller from feedback
& Adaptation \\

\makecell[l]{\texttt{/agent/}\\\texttt{run-autonomous-}\\\texttt{experiment}}
& Runs closed-loop execution with an executor
& Orchestration \\

\makecell[l]{\texttt{/experiments/}\\\texttt{\{run\_id\}}}
& Retrieves logged summaries and artifacts
& Reproducibility \\

\bottomrule
\end{tabular}
\end{table}
At round \(t\), the external executor submits diagnostics \(d_t\) according to the
registered schema \(\mathcal{S}\). The diagnostic adapter maps these named values
into the ordered state vector \(s_t=\phi_{\mathcal{S}}(d_t)\). The controller then
generates raw controls, which are projected into the schema-defined feasible
region before being returned to the executor given in Equation \ref{eq:bounded_control_suggestion}

The executor applies \(u_t\) in one classical, quantum-aware, hybrid, or
domain-specific optimization round, producing outputs \(z_t=E_t(u_t)\). These
outputs are mapped to feedback \(r_t=\psi(z_t)\), which is submitted to the API
for controller update and persistent logging. Therefore, the API does not require
direct access to raw data, model internals, aggregation logic, quantum circuits,
simulator state, or QPU job details.

Schema-bearing endpoints use fail-fast validation for missing diagnostics,
non-numeric diagnostics, unknown strategy identifiers, invalid control bounds,
missing controller identifiers, and incompatible schema versions. Control outputs
are bounded by schema-defined and user-defined limits before execution. This
policy supports safe reuse by external executors while preserving a traceable
record of diagnostics, controls, feedback, metrics, and artifacts.

Detailed JSON schema examples, versioned schema metadata, and structured endpoint
error responses, autonomous experiment requests, external-executor protocol
tables, round-level protocol pseudocode, and failure-handling policies are
provided in Supplementary Section~S4.


\section{Software Validation Methodology}
\label{sec:software_validation_methodology}

The software was validated using a layered methodology instead of a single demonstration run. The objective was to confirm that DeepUnfolding-Agent functions correctly at the component, API, closed-loop control, reproducibility, and deployment levels. This approach is well-suited for a software or API contribution, as the framework’s central claim is not only improved benchmark results, but also that the controller service can be reliably created, queried, updated, bounded, persisted, and reused by external executors.

The validation process includes five layers: unit, integration, system, artifact, and deployment validation. Unit validation ensures the correctness of individual components such as schemas, adapters, controller outputs, and safety bounds. Integration validation verifies that API endpoints and internal components interact as intended. System validation assesses closed-loop execution over multiple rounds. Artifact validation confirms that outputs needed for reproducibility are generated and recoverable. Deployment validation ensures the service can be launched and inspected outside the development environment.

\begin{table*}[t]
\centering
\scriptsize
\setlength{\tabcolsep}{2.2pt}
\renewcommand{\arraystretch}{1.08}
\caption{Automated software testing coverage and validation evidence. The optional IBM QPU credential-gated connection test is excluded from the default coverage suite.}
\label{tab:automated_testing_coverage}
\begin{tabularx}{\textwidth}{@{}p{0.14\textwidth}p{0.2\textwidth}X p{0.2\textwidth}p{0.13\textwidth}@{}}
\toprule
Validation layer & Tests or commands & Evidence checked & Coverage evidence & Result \\
\midrule

Unit validation
& \texttt{test\_controller\_core.py}, \texttt{test\_custom\_strategy.py}, \texttt{test\_torch\_controller.py}
& Schema parsing, bounds, diagnostics, feedback update, and checkpoint round-trip
& \texttt{torch\_controller}: 84\%; \texttt{sqlite\_store}: 83\%; \texttt{registry}: 92\%
& Pass \\

API integration
& \texttt{test\_api\_workflow.py}, \texttt{test\_agent\_route.py}
& Strategy registration, controller creation, control suggestion, update, and JSON responses
& \texttt{api/schemas}: 100\%; \texttt{api/main}: 52\%; routes: 100\%
& Pass \\

Safety validation
& \texttt{test\_agent\_safety\_guard.py}
& Bounded control clipping and type-safe classical/quantum controls
& \texttt{safety\_guard}: 90\%; \texttt{agent/schemas}: 100\%
& Pass \\

System validation
& \texttt{test\_agent\_orchestrator.py}, \texttt{test\_agent\_evaluator.py}
& Multi-round closed-loop execution, stopping logic, evaluation, and summaries
& \texttt{orchestrator}: 76\%; \texttt{evaluator}: 87\%; \texttt{simulated\_executor}: 100\%
& Pass \\

Quantum validation
& \texttt{test\_quantum\_api\_execution.py}
& QFL state mapping, bounded quantum controls, Qiskit Aer execution, shots, depth, and loss output
& \texttt{quantum\_wrappers}: 87\%; \texttt{quantum/backends}: 54\%
& Pass \\

Artifact validation
& Generated CSV, JSON, SQLite, checkpoints, and plots
& Round metrics, control traces, run configuration, summaries, and persisted controller state
& Functional artifact checks; checkpoint round-trip included
& Pass \\

Deployment validation
& \texttt{uvicorn}, Docker, \texttt{/health}, \texttt{/docs}
& Service startup, route availability, OpenAPI documentation, and containerised execution
& System-level validation; not counted in line coverage
& Pass \\

Full automated suite
& \texttt{pytest tests/ --cov=unfolded\_controller --cov-report=term-missing -ra}
& Default automated software validation suite
& \texttt{58\%} total line coverage over \texttt{1762} statements
& \texttt{21 pass, 3 skip} \\

\bottomrule
\end{tabularx}
\end{table*}

\begin{lstlisting}[language=bash,caption={Local software validation commands.},label={lst:software_validation_commands}]
python -c "from unfolded_controller.api.main import app; [print(sorted(r.methods), r.path) for r in app.routes]"
pytest tests/ -q
python -c "import requests; print(requests.get('http://127.0.0.1:8000/health').json())"
python -c "import requests; print(requests.get('http://127.0.0.1:8000/strategies').json())"
\end{lstlisting}

\begin{table}[ht]
\centering
\scriptsize
\setlength{\tabcolsep}{2.2pt}
\renewcommand{\arraystretch}{1.05}
\caption{Combined safety-bound validation results for classical and quantum controls.}
\label{tab:combined_safety_bound_validation}
\begin{tabular}{llllll}
\toprule
\makecell{Control\\family} 
& Control 
& \makecell{Unsafe\\input} 
& \makecell{Safety\\limit} 
& \makecell{Returned\\value} 
& \makecell{Limit\\source} \\
\midrule
\multirow{5}{*}{\rotatebox{90}{Quantum}}
& \texttt{theta\_lr} & 0.20 & 0.05 & 0.05 & User constraint \\
& \texttt{shots} & 9000 & 4096 & 4096 & User constraint \\
& \texttt{circuit\_depth} & 20 & 8 & 8 & User constraint \\
& \texttt{entangle\_depth} & 2 & 6 & 2 & Within bound \\
& \texttt{meas\_reuse} & 1.50 & 1.00 & 1.00 & Schema bound \\
\midrule
\multirow{3}{*}{\rotatebox{90}{Classical}}
& \texttt{lr} & 0.20 & 0.03 & 0.03 & User constraint \\
& \texttt{mu} & 0.50 & 0.20 & 0.20 & Schema bound \\
& \texttt{epochs} & 10 & 3 & 3 & User constraint \\
\bottomrule
\end{tabular}
\end{table}

\subsection{Unit Validation}
\label{subsec:unit_validation}

Unit validation verifies that the internal software components satisfy the assumptions required by the formal control contract. These tests cover request models, strategy schemas, control-variable bounds, diagnostic-to-state conversion, controller output types, safety-bound enforcement, feedback update behavior, and checkpoint round-trips. Schema tests confirm that each registered strategy contains a valid strategy identifier, controller family, ordered diagnostic variables, bounded control variables, and data types. Invalid schemas, missing diagnostics, inconsistent bounds, and non-numeric diagnostic values are expected to fail validation before controller inference.

TSafety-bound tests ensure that unsafe classical and quantum controls are not passed directly to the executor. Controls exceeding schema-defined or user-defined limits are adjusted to admissible values, while those within the valid range remain unchanged. This process verifies the bounded-control requirement of the proposed API. The safety guard ensures the controller does not apply controls outside schema-defined or user-defined bounds.

\subsection{Integration Validation}
\label{subsec:integration_validation}

Integration validation verifies the public API contract and the interaction between endpoint handlers, the strategy registry, controller factory, diagnostic adapter, controller backend, safety layer, and persistence store. The tested API workflow includes service availability, strategy discovery, user-defined strategy registration, controller creation, bounded control suggestion, feedback update, quantum-round execution, and experiment-output retrieval.

For each endpoint, the validation checks both the HTTP response status and the expected JSON response fields. This is important because external users interact with the system through the public HTTP interface rather than by directly instantiating internal Python classes. Successful endpoint validation confirms that an external executor can create a controller, submit diagnostics, receive bounded controls, return feedback, and retrieve logged results through the documented API boundary.

\subsection{System Validation}
\label{subsec:system_validation}

System validation determines whether DeepUnfolding-Agent supports repeated closed-loop operation instead of isolated endpoint calls. Each test covers the full lifecycle: controller creation, diagnostic submission, bounded control suggestion, external execution, feedback collection, controller update, and multiple repetitions. A successful test confirms that controller identifiers, diagnostics, controls, feedback metrics, and update records remain consistent throughout the process.

Autonomous orchestration tests assess the higher-level agent execution path.
These tests evaluate interactions among strategy selection, controller creation, safety enforcement, and executor invocation.
They also cover evaluator logic, stopping decisions, logging, and summary generation.
The oThe orchestration layer is valid if it completes bounded classical or quantum runs, generates a structured run summary, records safety-adjusted controls, and preserves the experiment trace.

In each round, diagnostics are mapped to a schema-defined state vector. The controller then proposes bounded adaptive controls, which the external executor applies. Feedback metrics are returned based on the execution backend.
The feedback is submitted to the agent controller update endpoint and logged in the experiment trace.

These tests are essential for validating the software’s adaptive behavior.
A successful closed-loop test confirms that controller identifiers, diagnostics, controls, feedback metrics, and update records remain consistent across multiple rounds.
It also verifies that the service supports a repeated feedback-control process, where runtime observations influence subsequent control suggestions, rather than returning static hyperparameters.

\subsection{Quantum Execution Validation}
\label{subsec:quantum_execution_validation}

Quantum execution validation determines whether the controller API supports quantum-aware workflows. The Qiskit Aer validation path ensures that quantum diagnostics, including quantum gradient norm, parameter-shift variance, shot noise, circuit depth, and round index, are correctly converted to the expected state representation. The controller then returns bounded quantum controls such as theta learning rate, number of shots, circuit depth, entanglement depth, and measurement reuse.

The quantum executor applies these bounded controls to a simulated quantum workflow and returns metrics including expectation value, training loss, validation loss, number of shots, number of qubits, and circuit depth. These metrics form the feedback signal for controller updates. Qiskit Aer serves as the default reproducible backend, enabling quantum workflow validation without requiring cloud credentials or physical hardware.

IBM QPU validation is considered optional hardware smoke testing rather than part of the default automated suite. Real-QPU execution depends on user credentials, network availability, queue status, and backend availability. Hardware validation confirms that credentials can be loaded, a non-simulator backend can be selected, a minimal job can be submitted, and backend or job metadata can be recorded.

\begin{table}[ht]
\centering
\scriptsize
\setlength{\tabcolsep}{3pt}
\caption{IBM QPU portability smoke-test result.}
\label{tab:ibm_qpu_portability_smoke_test}
\begin{tabular}{llll}
\toprule
Target & Evidence & Observed value & Result \\
\midrule
Credential loading & IBM account loaded & \texttt{ok} & Pass \\
Backend selection & Non-simulator backend selected & \texttt{ibm\_marrakesh} & Pass \\
Hardware check & Simulator flag verified & \texttt{false} & Pass \\
Job submission & QPU job submitted and completed & \texttt{completed} & Pass \\
Execution budget & Hardware shots used & 128 & Pass \\
Circuit size & Minimal circuit qubits & 1 & Pass \\
Compiled circuit & Transpiled circuit depth & 4 & Pass \\
\bottomrule
\end{tabular}
\end{table}

Table~\ref{tab:ibm_qpu_portability_smoke_test} reports an IBM QPU portability smoke test. 
This test verifies that the quantum execution abstraction can load IBM Quantum credentials, select a real hardware backend, submit a minimal job, complete execution, and record backend metadata.
This test serves solely as evidence of portability for hardware-backed execution support.
It does not evaluate optimization performance, accuracy improvement, latency, queueing behavior, or hardware superiority over simulation.

Qiskit Aer remains the default reproducible backend for automated testing because it does not depend on cloud credentials, queue availability, network access, or changing hardware conditions. 
The IBM QPU pathway is treated as an optional, credential-gated smoke test that demonstrates deployability to real quantum hardware when user access and backend availability allow.

\subsection{Artifact and Reproducibility Validation}
\label{subsec:Artefact _reproducibility_validation}

Artifact validation verifies that each run produces the outputs required for reproducibility and post-hoc inspection. These artifacts include round-level metrics, control histories, diagnostics, feedback records, configuration files, controller metadata, checkpoint files, and summary outputs. The purpose of this layer is to ensure that results are not only produced during execution but also stored in a form that can be inspected, compared, and reused.

This validation layer is especially important for an API-based controller because the evidence of correct behavior is distributed across diagnostics, controller decisions, executor metrics, feedback updates, and persisted metadata. A run is considered reproducible when its controller identifier, strategy identifier, schema, controls, feedback, and output artifacts can be traced after execution.

\subsection{Deployment Validation}
\label{subsec:deployment_validation}

Deployment validation ensures the service can operate outside the local development workflow. Local deployment is validated by launching the FastAPI service and verifying route registration, the health endpoint, and the OpenAPI/Swagger interface. Containerised deployment is validated by building the Docker image, launching the service, and confirming the API responds through the exposed port.

This deployment layer supports the framework’s external-user claim. It confirms that users can run the controller service without rebuilding the development environment and that external executors can access the service through a consistent HTTP interface in both local and containerised environments.

\begin{lstlisting}[language=bash,caption={Docker deployment validation commands.},label={lst:docker_deployment_validation}]
docker build -t deepunfolding-agent-api:0.1.0 .
docker run --rm -p 8000:8000 deepunfolding-agent-api:0.1.0
docker logs deepunfolding-agent-api-test
curl http://127.0.0.1:8000/health
curl http://127.0.0.1:8000/docs
\end{lstlisting}

\begin{table}[t]
\centering
\scriptsize
\setlength{\tabcolsep}{3pt}
\renewcommand{\arraystretch}{1.08}
\caption{Software validation environment and reproducibility metadata.}
\label{tab:software_validation_environment}
\begin{tabularx}{\columnwidth}{@{}p{0.32\columnwidth}Y@{}}
\toprule
\textbf{Item} & \textbf{Recorded value} \\
\midrule

OS/execution
& Windows 11 64-bit; Docker-based Linux container validation supported \\

Python/commit
& Python 3.14.6; Git commit \texttt{a80e040} \\

Docker image
& \texttt{deepunfolding-agent-api:0.1.0} \\

Docker image ID
& \texttt{\seqsplit{sha256:1987a3cf2b7be8e19eb0dd6cde856be469f57ef7e8e5ee7a401240e0a2498d15}} \\

Core packages
& FastAPI 0.138.0; PyTorch 2.12.1; Qiskit 2.4.2; Qiskit Aer 0.17.2 \\

Data / ML packages
& NumPy 2.5.0; Pandas 3.0.3; scikit-learn 1.9.0 \\

Test tools
& pytest 9.1.1; pytest-cov / coverage plugin 7.1.0 \\

Coverage command
& \texttt{pytest tests/ --cov=unfolded\_controller --cov-report=term-missing -ra} \\

Test outcome
& 21 passed, 3 skipped, 16 warnings in 63.32s \\

Line coverage
& 58\% total line coverage over 1762 executable statements \\

Hardware tests
& IBM QPU connection and execution tests are credential-gated and excluded from the default coverage suite \\

\bottomrule
\end{tabularx}
\end{table}
The complete validation command, optional IBM QPU smoke-test commands, and extended environment details are provided in Supplementary Section~S6.

\section{Reproducible External Usage and Evaluation}
\label{sec:external_usage_evaluation}

This section describes how an external user can reproduce the main execution pathways of the proposed software. The demonstrations include fixed-baseline and controller-adaptive runs. The fixed baseline uses manually specified controls, whereas the controlled run obtains adaptive controls from the deep-unfolded controller API. This enables the external user to compare whether the controller pathway executes correctly and whether it changes the training or execution trajectory relative to a static configuration.

The external-user benchmark uses datasets from financial fraud detection,
medical imaging, genomic bioinformatics, clinical diagnosis, and vehicle
trajectory analysis. This selection evaluates whether the proposed
diagnostic--control--feedback contract can be reused across heterogeneous
tabular and derived-feature workflows, rather than being tied to a single
dataset or domain. Detailed dataset sources, task roles, and references are
provided in Supplementary Section~S7.

\subsection{External User Setup}
\label{subsec:external_user_setup}

This subsection differentiates general API deployment from the DUAgent-specific external control protocol. Standard reproducibility mechanisms, such as starting the FastAPI service, inspecting the health endpoint, and deploying with Docker, establish the software boundary that allows DUAgent to provide deep-unfolded adaptive control to external classical, QFL, hybrid, and hardware-backed workflows.
The external-user package serves as the primary reproducibility artifact.
artifact: it provides the minimal commands, input format, expected outputs, and
extension points required for a first-time user to run DUAgent without modifying
The internal controller implementation.

An external user can start the DUAgent service from the local Python environment as follows:

\begin{lstlisting}[language=bash,caption={Starting the DUAgent API service locally.},label={lst:start_api_local}]
python -m uvicorn unfolded_controller.api.main:app --host 127.0.0.1 --port 8000 --reload
\end{lstlisting}

The running service can be inspected through the health endpoint and the Swagger documentation interface:

\begin{lstlisting}[language=bash,caption={Inspecting the running API service.},label={lst:inspect_api}]
curl http://127.0.0.1:8000/health
curl http://127.0.0.1:8000/docs
\end{lstlisting}

For containerised execution, the same service can be built and launched through Docker:

\begin{lstlisting}[language=bash,caption={Running the DUAgent API service through Docker.},label={lst:docker_external_usage}]
docker build -t deepunfolding-agent-api:0.1.0 .
docker run --rm -p 8000:8000 deepunfolding-agent-api:0.1.0
\end{lstlisting}

After either local or Docker deployment, the controller service is available through the same HTTP interface at \texttt{http://127.0.0.1:8000}. Docker therefore provides a reproducible packaging mechanism, while HTTP defines how external executors communicate with the running service.

\begin{table}[t]
\centering
\small
\caption{External usage modes supported by DUAgent.}
\label{tab:external_usage_modes}
\begin{tabular}{p{0.22\linewidth}p{0.32\linewidth}p{0.33\linewidth}}
\toprule
Usage mode & Purpose & User interaction \\
\midrule
Repository examples
& Reproduce the reported benchmark workflows
& Start the API and run the provided external executor scripts \\

Docker deployment
& Package and run DUAgent as a containerised service
& Expose the API on port \texttt{8000} and communicate with it through HTTP \\

External HTTP executor
& Connect an independent training, optimization, simulator, or QPU workflow
& Send diagnostics, receive bounded controls, execute locally, and return feedback \\
\bottomrule
\end{tabular}
\end{table}

\subsection{Artifact  Reproducibility Checklist}
\label{subsec:Artefact _reproducibility_checklist}

To make the external-user artifact reproducible, the repository includes a dedicated
\texttt{external/} entry point containing a quick-start CSV generator, a public
external runner, strategy schemas, executor templates, and usage documentation.
The checklist in Table~\ref{tab:artifact_reproducibility_checklist} summarizes the minimum steps required to install the software, start the controller service, run an external workflow, and inspect the generated artifacts.

\begin{table}[htb]
\centering
\scriptsize
\setlength{\tabcolsep}{3pt}
\renewcommand{\arraystretch}{1.08}
\caption{Compact artifact reproducibility checklist for external users.}
\label{tab:artifact_reproducibility_checklist}
\begin{tabularx}{\columnwidth}{@{}p{0.31\columnwidth}Y@{}}
\toprule
\textbf{Checklist item} & \textbf{Expected reproducibility evidence} \\
\midrule

Environment setup
& Python or Docker environment created with repository dependencies installed. \\

API service validation
& DUAgent service starts successfully; \texttt{/health} and \texttt{/docs} confirm API availability. \\

External workflow execution
& Fixed, heuristic, and DUAgent-controlled workflows execute through the public API using user-provided CSV data. \\

Round-level logging
& \texttt{round\_metrics.csv} records diagnostics, bounded controls, losses, scores, and execution metadata. \\

Final summaries
& \texttt{final\_summary.csv} and \texttt{run\_config.json} record final metrics, task configuration, seed, rounds, clients, strategy, and output path. \\

Partition and plot evidence
& Client label distributions and generated learning-trajectory plots support post-hoc reproducibility analysis. \\

Custom strategy extension
& User-defined schemas and executor plugins can be tested without modifying API internals. \\

\bottomrule
\end{tabularx}
\end{table}

This checklist distinguishes Artifact reproducibility from benchmark performance.
A successful reproduction does not require the user to reproduce every reported
Instead, the minimum Artifact requirement is that users can
install the software, start the API, run the external CSV workflow,
obtain bounded DUAgent controls through the public service, and generate
the expected logs, CSV summaries, configuration files, and plots. The same interface
can then be extended to user-defined classical or quantum workflows by replacing
the external schema and executor plugin while keeping the API unchanged.

The complete step-by-step reproducibility checklist, including installation commands,
API start-up commands, sample-data generation, output-file inspection, and custom
strategy extension steps, is provided in Supplementary Section~S8.

\subsection{External HTTP Control Protocol}
\label{subsec:http_executor_usage}

DUAgent provides a closed-loop control protocol over HTTP. In this setup, the external executor manages the dataset, preprocessing pipeline, model, simulator, or QPU backend. DUAgent manages the adaptive controller and offers controller creation, control suggestions, and feedback updates through public API endpoints. As a result, the external workflow interacts with DUAgent as an adaptive-control service rather than instantiating the controller directly.

The protocol includes three service interactions. First, the executor creates a controller using either a built-in or user-registered strategy. Second, during each round, it sends runtime diagnostics and receives bounded control variables. Third, after applying these controls, it returns observed feedback for controller updates. This interaction pattern applies to classical, QFL, hybrid, and real-QPU experiments.
Table~\ref{tab:http_executor_usage} summarizes the execution stages. The controller is domain-agnostic and does not manage the external dataset, training loop, simulator, or QPU job submission. Instead, it exchanges diagnostics, controls, and feedback with the executor in each round.

\begin{table}[t]
\centering
\small
\caption{External HTTP control protocol.}
\label{tab:http_executor_usage}
\begin{tabular}{p{0.05\linewidth}p{0.40\linewidth}p{0.38\linewidth}}
\toprule
Step & External executor action & DUAgent API role \\
\midrule
1 & Requests a controller for a selected strategy & Creates a reusable controller instance \\
2 & Sends runtime diagnostics & Returns bounded adaptive controls \\
3 & Applies controls to the local workflow & Remains independent of domain-specific execution \\
4 & Sends validation loss and task feedback & Updates the controller state \\
5 & Saves round-level outputs & Supports reproducible evaluation through logs and summaries \\
\bottomrule
\end{tabular}
\end{table}

\begin{figure*}[htb]
    \centering
    \includegraphics[width=0.95\linewidth]{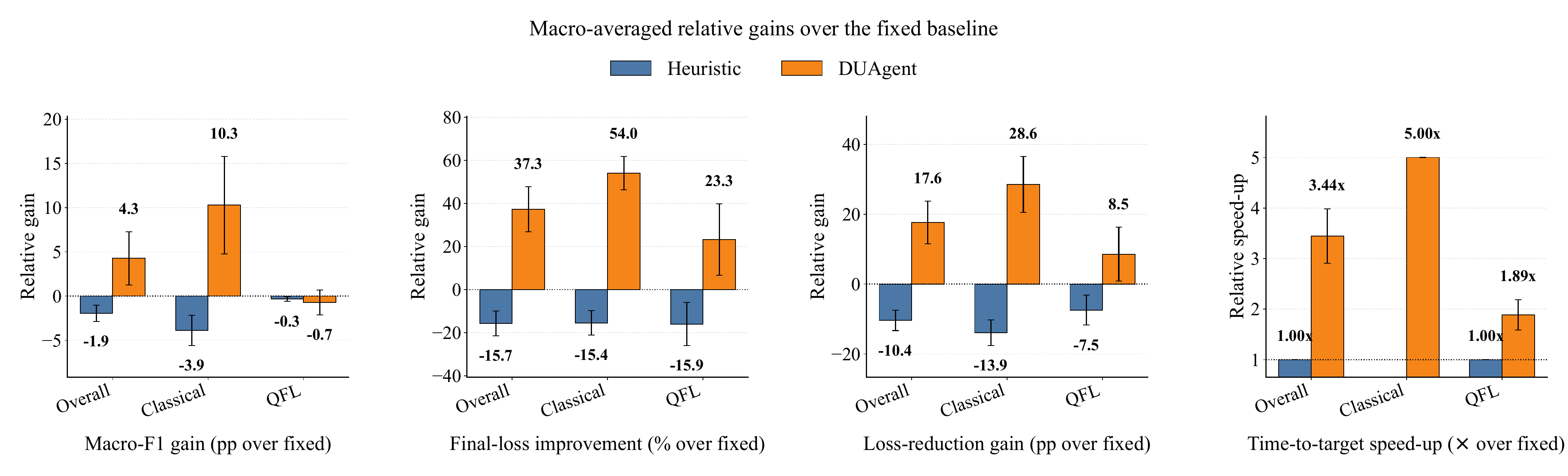}
    \caption{
    Macro-averaged relative gains over the fixed baseline. 
    Each dataset-workflow pair is first normalized against its own fixed baseline before aggregation.
    }
    \label{fig:duagent_aggregate_relative_bar_summary}
\end{figure*}

The repository includes benchmark executors to support reproducibility. These executors use the same HTTP pathway as independent user code. Listing~\ref{lst:http_executor_client} presents the client-side interface for the external benchmark executor. The executor retrieves the service address from \texttt{CONTROLLER\_API\_BASE\_URL}, sends HTTP requests with \texttt{requests.get} and \texttt{requests.post}, and creates controllers through the public \texttt{/controllers/create} endpoint.

\begin{lstlisting}[language=Python,caption={HTTP client interface used by the external benchmark executor.},label={lst:http_executor_client}]
BASE_URL = os.getenv(
    "CONTROLLER_API_BASE_URL",
    "http://127.0.0.1:8000"
)

def api_get(path: str) -> dict:
    response = requests.get(f"{BASE_URL}{path}")
    response.raise_for_status()
    return response.json()

def api_post(path: str, payload: dict) -> dict:
    response = requests.post(f"{BASE_URL}{path}", json=payload)
    response.raise_for_status()
    return response.json()

response = requests.post(
    f"{BASE_URL}/controllers/create",
    json=payload
)
\end{lstlisting}

Both the classical and QFL benchmark runs evaluate the same public service interface used by external systems. The benchmark executor provides diagnostics, DUAgent returns bounded controls, the executor completes a workflow round, and the resulting feedback is sent to the API for the next controller update.

This protocol can be executed programmaticaly using the benchmark executor or manually via the public HTTP endpoints. In these experiments, the benchmark executor used the programmatic HTTP client interface shown in Listing~\ref{lst:http_executor_client}.

\subsection{Classical and Quantum FL External Benchmark Execution}
\label{subsec:external_benchmark_execution}

After starting the DeepUnfolding-Agent service, the same external-executor pattern is used for both classical FL and QFL workflows. The executor loads a user-provided CSV dataset, prepares the learning task, requests bounded controls from the DUAgent API, executes one workflow round, and submits validation feedback for controller update. Round-level outputs are stored as reproducibility artefacts and used to generate aggregate, per-dataset, and per-workflow results.

In the classical workflow, returned controls are applied to learning and local-training hyperparameters, such as learning rate, proximal strength, and local epochs. In the QFL workflow, returned controls are applied to quantum-relevant variables, including theta learning rate, shot count, circuit depth, entanglement depth, and measurement reuse. The executor remains responsible for domain-specific data loading, training, and evaluation, while DUAgent provides reusable adaptive control through the same diagnostic--control--feedback service interface.

The full external benchmark command template and dataset-specific execution examples are provided in Supplementary Section~S7.

\begin{table}[t]
\centering
\scriptsize
\setlength{\tabcolsep}{5.2pt}
\renewcommand{\arraystretch}{1.05}
\caption {Main external-user benchmark results. F1 and Red. are percentages; Spd. is speed-up relative to the fixed baseline.}
\label{tab:duagent_main_external_results}
\begin{tabular}{@{}lcccc@{}}
\toprule
\textbf{Method} & \textbf{F1$\uparrow$} & \textbf{Loss$\downarrow$} & \textbf{Red.$\uparrow$} & \textbf{Spd.$\uparrow$} \\
\midrule

\multicolumn{5}{@{}l}{\textbf{Overall}} \\
Fixed     & 48.9 & 0.658 & 24.2 & 1.00$\times$ \\
Heuristic & 47.0 & 0.715 & 13.8 & 1.00$\times$ \\
DUAgent   & \textbf{53.2} & \textbf{0.484} & \textbf{41.8} & \textbf{3.44$\times$} \\

\midrule
\multicolumn{5}{@{}l}{\textbf{Classical}} \\
Fixed     & 62.6 & 0.613 & 28.1 & 1.00$\times$ \\
Heuristic & 58.8 & 0.685 & 14.1 & NR \\
DUAgent   & \textbf{72.9} & \textbf{0.304} & \textbf{56.6} & \textbf{5.00$\times$} \\

\midrule
\multicolumn{5}{@{}l}{\textbf{QFL}} \\
Fixed     & \textbf{37.5} & 0.695 & 21.0 & 1.00$\times$ \\
Heuristic & 37.2 & 0.740 & 13.5 & 1.00$\times$ \\
DUAgent   & 36.8 & \textbf{0.634} & \textbf{29.5} & \textbf{1.89$\times$} \\

\bottomrule
\end{tabular}
\end{table}

In this template, \texttt{--dataset-name} identifies the benchmark dataset, \texttt{--mode} selects either the classical or QFL workflow, \texttt{--csv} specifies the user-provided tabular input file, and \texttt{--label} identifies the target column. 
The remaining arguments specify the number of communication rounds, the number of clients, the non-IID partitioning strength, the random seed, and the output directory. 
This template was applied to healthcare, finance, transport, and bioinformatics datasets by modifying only the dataset-specific arguments.

\section{Per-Dataset External Benchmark Results}
\label{app:per_dataset_results}

\begin{table}[htbp]
\centering
\scriptsize
\setlength{\tabcolsep}{3.2pt}
\renewcommand{\arraystretch}{1.05}
\caption{Compact external-user benchmark summary comparing fixed execution and DUAgent-controlled execution.}
\label{tab:duagent_external_results_compact_main}
\begin{threeparttable}
\begin{tabular}{@{}llrrrrrrr@{}}
\toprule
\textbf{WF} & \textbf{Data} 
& \makecell{\textbf{F1}\\\textbf{Fix.}} 
& \makecell{\textbf{F1}\\\textbf{DU}} 
& \makecell{\(\boldsymbol{\Delta}\)\textbf{F1}\\\textbf{pp}} 
& \makecell{\textbf{Loss}\\\textbf{Fix.}} 
& \makecell{\textbf{Loss}\\\textbf{DU}} 
& \makecell{\textbf{Red.}\\\textbf{DU}} 
& \makecell{\textbf{Spd.}\\\textbf{DU}} \\
\midrule

C & BAF   & 40.4 & 49.7  & +9.3  & 0.630 & 0.231 & 64.9 & 5.00 \\
C & BL    & 76.0 & 81.9  & +5.9  & 0.524 & 0.373 & 32.8 & 5.00 \\
C & Hanta & 51.5 & 83.2  & +31.7 & 1.209 & 0.656 & 49.1 & 5.00 \\
C & NGSIM & 48.1 & 49.8  & +1.7  & 0.498 & 0.207 & 59.9 & 5.00 \\
C & WBC   & 97.2 & 100.0 & +2.8  & 0.205 & 0.053 & 76.4 & 5.00 \\

\midrule

Q & BAF   & 49.7 & 49.7 & 0.0  & 0.313 & 0.070 & 88.8 & 2.00 \\
Q & BL    & 31.7 & 28.2 & -3.5 & 0.708 & 0.712 & -0.9 & 1.00 \\
Q & Hanta & 9.7  & 14.3 & +4.6 & 1.556 & 1.524 & 8.2  & 1.43 \\
Q & HDSVT & 45.5 & 40.2 & -5.3 & 0.693 & 0.694 & 0.1  & 2.50 \\
Q & NGSIM & 49.8 & 49.8 & 0.0  & 0.247 & 0.067 & 87.5 & 2.50 \\
Q & WBC   & 38.7 & 38.7 & 0.0  & 0.655 & 0.735 & -6.5 & NR \\

\bottomrule
\end{tabular}
\begin{tablenotes}
\scriptsize
\item \textit{Note:} C denotes the classical workflow and Q denotes the QFL workflow. BL denotes Breast Lesions, Hanta denotes Hantavirus, and WBC denotes Wisconsin Breast Cancer. F1 and Red. are percentages. \(\Delta\)F1 is the DUAgent--Fixed difference in percentage points. Loss is final validation loss. Spd. is time-to-target speed-up relative to the fixed baseline; NR indicates that the target was not reached. Full fixed, heuristic, and DUAgent results are provided in Supplementary Section~S8.
\end{tablenotes}
\end{threeparttable}
\end{table}

Figure~\ref{fig:duagent_aggregate_relative_bar_summary} reports paired relative gains over the fixed baseline. 
DUAgent delivers the largest gains in the classical workflow, increasing Macro-F1 by 10.3 percentage points, reducing final loss by 54.0\%, improving loss reduction by 28.6 percentage points, and accelerating time-to-target by a factor of 5.00.
Across all workflows, DUAgent improves final loss by 37.3\% and reaches the target 3.44$\times$ faster than the fixed baseline. 
The QFL workflow shows more moderate but still useful optimization gains: DUAgent improves final loss by 23.3\%, increases loss reduction by 8.5 percentage points, and reaches the target 1.89$\times$ faster. 
However, QFL Macro-F1 does not consistently improve, suggesting that the current controller mainly improves QFL convergence behavior rather than uniformly improving final predictive performance.

Table~\ref{tab:duagent_main_external_results} presents an aggregate comparison of fixed, heuristic, and DUAgent-controlled execution across external benchmark workflows.
The results show that DUAgent achieves the best overall final Macro-F1, final validation loss, loss reduction, and time-to-target speed-up. 
The most significant improvements occur in the classical workflow, where DUAgent increases final Macro-F1 from 62.6\% to 72.9\%, reduces final validation loss from 0.613 to 0.304, and achieves the target 5.00$\times$ faster than the fixed baseline.
For QFL, DUAgent delivers the best final validation loss and loss reduction, while final Macro-F1 remains similar to the fixed and heuristic baselines.
This suggests that the current QFL controller enhances optimization and convergence speed more than final predictive performance.

Supplementary Table S8  reports final task performance, final validation loss, loss reduction, and time-to-target speed-up for each dataset and workflow. 
Speed-up is reported relative to the fixed baseline, where 1.00$\times$ denotes parity, and bold values indicate the best result within each dataset and workflow.

\subsection{Quantum Qiskit Aer Usage}
\label{subsec:qiskit_aer_usage}
The quantum API pathway presents quantum execution as a controllable, schema-driven workflow instead of a fixed experiment script.
In this framework, the external executor sends quantum diagnostics to the DeepUnfolding-Agent service, receives bounded quantum-aware controls, executes the quantum round, and returns feedback metrics for controller updates.
This design is essential because quantum workflows introduce control variables not present in classical learning, such as shot count, circuit depth, entanglement depth, parameter-update scale, measurement reuse, and noise-sensitive diagnostic signals.
Representing these variables through the same public API contract as the classical workflow enables a reusable interface for adaptive quantum control.

Qiskit Aer is used as the default reproducible quantum backend. 
This pathway enables testing of the full quantum control loop without cloud credentials, queue availability, or physical quantum hardware.
External users can execute a controlled QFL round through the public API endpoint or the provided demonstration script.
In both cases, quantum diagnostics are mapped to a controller state vector, bounded QFL controls are generated, and the executor returns quantum execution metrics, including expectation value, validation loss, number of shots, number of qubits, and circuit depth.

The endpoint-based quantum execution test was successful, confirming that the \texttt{POST /quantum/run-controlled-round} endpoint can be accessed through the public API using the Qiskit Aer backend.
This result validates the primary role of the quantum API: reproducible simulator-based execution with adaptive controller interaction.

\begin{figure*}[htb]
    \centering
    \includegraphics[width=0.95\linewidth]{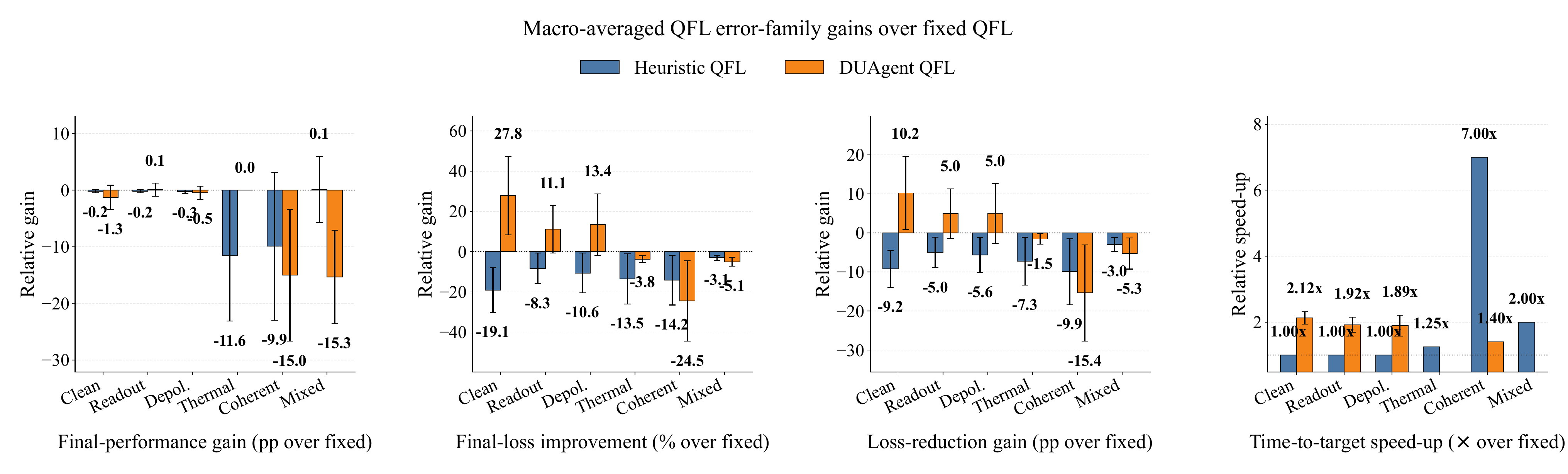}
\caption{
Macro-averaged QFL error-family gains over the fixed-QFL baseline.
The figure reports a noise-family stress test rather than a uniform performance claim.
Bars show relative gains after normalising each dataset--error-family result against its own fixed-QFL baseline.
Error bars denote the standard error of the mean across dataset observations, not across random seeds.
The mixed behavior across noise families identifies where the current controller is effective and where additional noise-aware diagnostics or update rules are needed.}
\label{fig:aggregate_relative_gains}
    \label{fig:qfl_error_family_aggregate_relative_summary}
\end{figure*}

Figure~\ref{fig:aggregate_relative_gains} presents macro-averaged relative gains compared to the fixed baseline.
The aggregation incorporates three datasets: BAF, Hantavirus, and NGSIM, each evaluated under FL/QFL workflow settings.
For each dataset and workflow combination, the results for both the Heuristic and DUAgent methods are normalized against the fixed baseline from the same dataset and workflow.
Consequently, the plotted bars represent relative gains following per-task normalization, rather than raw metric values aggregated across heterogeneous datasets.

The error bars indicate the standard error of the mean calculated across the normalized dataset and workflow observations within each aggregation group.
Because each experiment was conducted with a single random seed per dataset and workflow setting, the error bars reflect variation across datasets and workflow settings, rather than uncertainty from repeated random seeds.
Overall, DUAgent demonstrates positive aggregate gains over the fixed baseline for Macro-F1, final-loss improvement, loss reduction, and time-to-target speed-up. In contrast, the heuristic baseline exhibits less consistent performance and can yield negative relative gains in several settings. 

\begin{table}[hbt]
\centering
\caption{
QFL behavior under different quantum error families. Values are macro-averaged across datasets and reported relative to the fixed-QFL baseline for each dataset--error-family pair. The best value within each noise condition and metric is shown in bold.
}
\label{tab:qfl_error_family_method_behavior}
\scriptsize
\setlength{\tabcolsep}{2.5pt}
\renewcommand{\arraystretch}{1.08}

\begin{tabular}{llrrrr}
\toprule
Noise & Method & $\Delta$F1$\uparrow$ & $\Delta$Loss$\uparrow$ & $\Delta$Red.$\uparrow$ & Speed$\uparrow$ \\
      &        & (pp)                 & (\%)                   & (pp)                   & ($\times$) \\
\midrule

\multirow{3}{*}{Clean}
& Fixed    & \textbf{0.0}  & 0.0            & 0.0            & 1.00 \\
& Heur.    & -0.2          & -19.1          & -9.2           & 1.00 \\
& DUAgent  & -1.3          & \textbf{27.8}  & \textbf{10.2}  & \textbf{2.12} \\

\midrule
\multirow{3}{*}{Readout}
& Fixed    & 0.0           & 0.0            & 0.0            & 1.00 \\
& Heur.    & -0.2          & -8.3           & -5.0           & 1.00 \\
& DUAgent  & \textbf{0.1}  & \textbf{11.1}  & \textbf{5.0}   & \textbf{1.92} \\

\midrule
\multirow{3}{*}{Depol.}
& Fixed    & \textbf{0.0}  & 0.0            & 0.0            & 1.00 \\
& Heur.    & -0.3          & -10.6          & -5.6           & 1.00 \\
& DUAgent  & -0.5          & \textbf{13.4}  & \textbf{5.0}   & \textbf{1.89} \\

\midrule
\multirow{3}{*}{Thermal}
& Fixed    & \textbf{0.0}  & \textbf{0.0}   & \textbf{0.0}   & 1.00 \\
& Heur.    & -11.6         & -13.5          & -7.3           & \textbf{1.25} \\
& DUAgent  & \textbf{0.0}  & -3.8           & -1.5           & -- \\

\midrule
\multirow{3}{*}{Coherent}
& Fixed    & \textbf{0.0}  & \textbf{0.0}   & \textbf{0.0}   & 1.00 \\
& Heur.    & -9.9          & -14.2          & -9.9           & \textbf{7.00} \\
& DUAgent  & -15.0         & -24.5          & -15.4          & 1.40 \\

\midrule
\multirow{3}{*}{Mixed}
& Fixed    & 0.0           & \textbf{0.0}   & \textbf{0.0}   & 1.00 \\
& Heur.    & \textbf{0.1}  & -3.1           & -3.0           & \textbf{2.00} \\
& DUAgent  & -15.3         & -5.1           & -5.3           & -- \\

\bottomrule
\end{tabular}%
\end{table}

The Qiskit Aer pathway provides a reproducible simulator-based quantum execution route. This pathway is used as the default quantum demonstration because it does not require cloud credentials or queue access. The external user can execute a controlled quantum round through the public API or through the provided quantum demonstration script. In both cases, quantum diagnostics are mapped into the controller state vector, bounded quantum-aware controls are generated, and the quantum executor returns execution metrics.

The endpoint-based quantum execution was validated successfully, with the test passing and no execution failures. This confirms that the \texttt{POST /quantum/run-controlled-round} endpoint can be invoked through the public API using the Qiskit Aer backend.

\subsection{Quantum Error-Family Evaluation}
\label{subsec:quantum_error_family_evaluation}

The second role of the quantum API is to support noise-aware evaluation. 
Real quantum execution is affected by readout error, gate noise, decoherence, coherent miscalibration, shot noise, and mixed hardware effects. 
While Qiskit Aer cannot replace real hardware validation, it offers a controlled simulation environment where various quantum error families can be reproducibly injected and studied.
This allows the proposed API to be evaluated under both ideal and noisy quantum conditions.

Accordingly, this experiment serves as a noise-family stress test for the adaptive-control interface.
The goal is not to demonstrate DUAgent's superiority in all noise conditions, but to identify where the current controller is effective, neutral, or inadequate under structured quantum noise.

To assess this capability, QFL experiments were conducted across multiple error families: clean execution, readout noise, depolarising noise, thermal relaxation, coherent error, and mixed noise.
For each error family, fixed QFL, heuristic QFL, and DUAgent QFL were evaluated using a consistent external executor protocol.
This experiment does not aim to claim quantum advantage, but rather to evaluate whether the DUAgent control interface can adapt quantum-relevant controls under noisy simulation conditions.
The executor reports quantum diagnostics to the API, receives bounded QFL controls, executes the noisy quantum round, and returns validation loss and task metrics for controller updates.

Figure~\ref{fig:qfl_error_family_aggregate_relative_summary} summarizes the relative behavior of heuristic QFL and DUAgent QFL under different quantum error families. 
The results are intentionally varied.
DUAgent delivers meaningful optimization gains under clean, readout-error, and depolarising-noise conditions, particularly in final-loss improvement, loss-reduction, and time-to-target acceleration.
This demonstrates that the proposed API can expose quantum diagnostics and use them to adapt control variables influencing QFL optimization when noise is absent, measurement-dominated, or approximately stochastic.

However, coherent and mixed noise present ongoing challenges.
In these scenarios, DUAgent does not consistently outperform the fixed or heuristic QFL baselines.
This suggests that the current diagnostic set is insufficient to fully characterize structured unitary miscalibration or compound hardware-like noise effects.
Unlike readout or depolarising noise, coherent and mixed noise may require additional calibration-aware diagnostics, noise-family identifiers, circuit-level error summaries, backend-specific metadata, or noise-specific feedback.

Table~\ref{tab:qfl_error_family_method_behavior} provides the corresponding numerical summary. 
DUAgent achieves the strongest optimization under clean, readout, and depolarizing conditions, while its performance under thermal, coherent, and mixed noise is weaker or less consistent.
This is scientifically valuable because it demonstrates that the API does not simply produce optimistic aggregate results.
Instead, the same diagnostic-control-feedback interface reveals both the regimes where adaptive control is effective and those where the current controller design requires further noise-aware modeling.

Representative classical and QFL trajectory plots are provided in Supplementary Section~S9.
Detailed validation-loss trajectories and DUAgent control responses for each noise family are provided in Supplementary Section~S12

\subsection{Real IBM QPU Execution}
\label{subsec:real_ibm_qpu_execution}

The third role of the quantum API is to support hardware-oriented execution.
The same external control protocol used for Qiskit Aer and noisy simulator
experiments can also route feature-encoded circuits to IBM Quantum hardware.
In this pathway, the executor loads a user-provided CSV dataset, constructs a
low-dimensional quantum representation, requests bounded QFL controls from
DUAgent, submits circuits to the selected IBM backend, and returns validation
metrics for controller update.

This experiment validates API portability, backend integration, and closed-loop
feedback on real quantum hardware; it is not intended as evidence of quantum
advantage or performance superiority over simulation. The controller does not
require knowledge of whether execution is performed by Qiskit Aer or an IBM QPU
backend. Backend-specific operations remain inside the executor, while the API
contract remains unchanged: diagnostics are submitted, bounded controls are
returned, execution is performed externally, and feedback is returned for
controller update.

\begin{table}[htbp]
\centering
\scriptsize
\caption{Compact real IBM QPU execution summary. The purpose is to validate hardware-backed API portability, not quantum advantage.}
\label{tab:compact_real_ibm_qpu}
\begin{tabular}{llcccc}
\toprule
Dataset & Method & Backend & Rounds & Shots & Macro-F1 \\
\midrule
NGSIM & Fixed QFL & \texttt{ibm\_marrakesh} & 10 & 1024 & 0.496 \\
NGSIM & DUAgent QFL & \texttt{ibm\_marrakesh} & 10 & 1024 & 0.496 \\
Hantavirus & Fixed QFL & \texttt{ibm\_marrakesh} & 10 & 1024 & 0.333 \\
Hantavirus & DUAgent QFL & \texttt{ibm\_marrakesh} & 10 & 1024 & 0.333 \\
\bottomrule
\end{tabular}
\end{table}

The hardware-backed runs demonstrate that DUAgent can maintain the same
diagnostic--control--feedback loop under real backend constraints. Detailed IBM
account setup commands, optional connection tests, execution templates,
fixed-versus-DUAgent hardware results, and learnability trajectories are
provided in Supplementary Section~S13.

\section{Limitations}
\label{sec:limitations}

This section summarizes the main limitations of the current DUAgent software artifact and evaluation. These limitations clarify the boundary between a validated research prototype and a production-ready adaptive-control system.

First, the evaluation focuses on software-level validation and reproducible benchmark execution rather than large-scale deployment. Experiments use limited dataset sizes, communication rounds, seeds, and execution budgets to maintain practical runtimes in local and Docker-based environments. Larger-scale studies with more seeds, longer runs, and stronger task-specific baselines are required before making broad performance claims.

Second, the benchmark datasets and preprocessing pipelines are selected to demonstrate external-user integration across healthcare, finance, genomics, transport, classical FL, QFL, and noisy-QFL workflows. Although this provides evidence of API-level reusability, it does not establish domain-general superiority. Future work should evaluate DUAgent with larger datasets, domain-specific models, and more comprehensive baseline comparisons.

Third, the QFL and IBM QPU experiments should be interpreted as quantum-aware control and hardware-portability evidence, not as evidence of quantum advantage. The simulator-based QFL workflows use reduced feature sets, limited circuit depths, constrained shot budgets, and simplified variational models. Real IBM QPU execution is also affected by queue delays, calibration drift, device noise, backend availability, and shot limits.

Fourth, the controller returns bounded control suggestions based on selected diagnostics and feedback signals, but it does not guarantee globally optimal policies. Performance depends on the diagnostic variables, control bounds, feedback objective, controller configuration, safety constraints, and executor backend. A more detailed component-wise ablation study is required to isolate the effects of individual diagnostics, controls, unfolding depth, controller capacity, feedback terms, and projection bounds.

Finally, DUAgent is designed as a research-oriented adaptive-control API, not as a complete secure production platform. The current safety guard constrains returned control values, but production deployment would require authentication, authorization, encrypted transport, rate limiting, secure credential management, audit logging, monitoring, fault recovery, privacy-preserving handling, and protection against adversarial diagnostics or poisoned feedback. Extended ablation and security limitations are provided in Supplementary Section~S14.


\section{Conclusion}
\label{sec:conclusion}

This paper introduced DeepUnfolding-Agent, a strategy-agnostic adaptive-control API for classical, quantum-aware, and hybrid learning workflows. The framework exposes deep-unfolded adaptive control through a reusable FastAPI service, allowing external users to register strategy schemas, create controllers, request bounded control suggestions, update controllers from feedback, and retrieve reproducible experiment artefacts. By separating controller intelligence from task-specific execution, DUAgent enables the same diagnostic--control--feedback interface to be reused across heterogeneous optimisation workflows.

The evaluation demonstrates the main software contribution: DUAgent provides a reusable, bounded, and externally executable adaptive-control layer. Classical experiments show that the API can regulate controls such as learning rate, proximal strength, and local epochs across multiple benchmark tasks. Quantum-aware experiments show that the same interface can control quantum-relevant variables, including theta learning rate, shot count, circuit depth, and measurement reuse. These results support the feasibility of a unified controller-agent interface for classical and quantum-aware adaptive optimisation.

Future work will extend DUAgent through larger multi-seed evaluations, stronger domain-specific baselines, more expressive quantum models, broader hardware-backed execution, and production-oriented features such as secure authentication, multi-user tracking, persistent controller management, and integration with privacy-preserving federated workflows. Overall, DUAgent establishes a reusable software foundation for adaptive-agent services that can support current classical learning systems and emerging quantum-aware machine-learning pipelines.

\section*{Software Availability}
\label{sec:software_availability}

DeepUnfolding-Agent is released as an open-source research software artifact. 
The public project website provides documentation, overview material, and links to the implementation repository. 
The repository contains the FastAPI service, controller backends, strategy schemas, external-user runners, benchmark scripts, Docker configuration, automated tests, and reproducibility artifacts.

\textbf{Project website:} \url{https://deepunfolding-agent.github.io/}

\textbf{Source code:} \url{https://github.com/shanikairoshi/DeepUnfolding-Agent}

\bibliographystyle{elsarticle-num}
\bibliography{references}

\end{document}